\documentclass[10pt,leqno]{amsart}
\usepackage{graphicx}
\usepackage{indentfirst,csquotes}
\usepackage[numbers]{natbib}

\usepackage{amssymb,amsthm,amsmath}
\usepackage{xcolor,paralist,hyperref,titlesec,fancyhdr,etoolbox}

\makeatletter
\let\uppercasenonmath\@gobble

\makeatother

\titleformat{\section}
  {\normalfont\Large\bfseries}
  {\thesection.}{0.5em}{}

\titlespacing*{\section}
  {0pt}{1.5ex plus .5ex minus .2ex}{0.8ex plus .2ex}

\titleformat{\subsection}
  {\normalfont\large\bfseries}
  {\thesubsection.}{0.5em}{}

\titlespacing*{\subsection}
  {0pt}{1.25ex plus .5ex minus .2ex}{0.7ex plus .2ex}

\titleformat{\subsubsection}
  {\normalfont\normalsize\bfseries}
  {\thesubsubsection.}{0.5em}{}

\titlespacing*{\subsubsection}
  {0pt}{1ex plus .5ex minus .2ex}{0.5ex plus .2ex}

\hypersetup{
  colorlinks=true,
  linkcolor=black,
  filecolor=black,
  urlcolor=black
}

\newcommand{\BigO}{\mathcal{O}}
\newcommand{\defeq}{\overset{\mathrm{def}}{=\joinrel=}}

\begin{document}

\title{Solving Convex Partition Visual Jigsaw Puzzles}
\author{
Yaniv Ohayon$^{1}$, Ofir Itzhak Shahar$^{1}$, and Ohad Ben-Shahar$^{1}$\\[4pt]
$^{1}$Ben-Gurion University of the Negev, Beer-Sheva, Israel\\[2pt]
\texttt{yanivoha@post.bgu.ac.il, shofir@post.bgu.ac.il, obs@bgu.ac.il}
}

\maketitle


\begin{abstract}
Jigsaw puzzle solving requires the rearrangement of unordered pieces into their original pose in order to reconstruct a coherent whole, often an image, and is known to be an intractable problem. While the possible impact of automatic puzzle solvers can be disruptive in various application domains, most of the literature has focused on developing solvers for \textit{square} jigsaw puzzles, severely limiting their practical use. %
In this work, we significantly expand the types of puzzles handled computationally, focusing on what is known as Convex Partitions, a major subset of polygonal puzzles whose pieces are convex. We utilize both geometrical and pictorial compatibilities, introduce a greedy solver, and report several performance measures next to the first benchmark dataset of such puzzles.
\end{abstract}

\bigskip

\section{Introduction}

While solving jigsaw puzzles is often seen as a leisure time activity, developing algorithmic approaches to this problem has meaningful applications in fields like archaeology~\cite{brown2008system}, biology~\cite{marande2007mitochondrial}, and shredded document restoration~\cite{jalkanen2017semi}, to name but a few. 
Generally speaking, solving a jigsaw puzzle involves reordering and reassembling unsordered pieces to reconstruct a coherent whole, often an image. This process relies on matching shapes of pieces (or fragments) and, in the case of pictorial puzzles, also using and 
aligning their visual content. However, as the problem is NP-complete~\cite{demaine2007jigsaw}, heuristics are essential to manage the combinatorial complexity.

The pioneering computational attempt to solve a jigsaw puzzle was made in the 1960s by Freeman and Garder~\cite{freeman1964apictorial}, who addressed geometric puzzles and experimented with small puzzles of just few pieces whose shapes were virtually unrestricted and devoid of any pictorial content. 
However, as research on the topic evolved, the critical mass of the literature on jigsaw puzzle solving has focused mainly on jigsaw puzzles with square pieces, known as \textit{square jigsaw puzzles}~\cite{toyama2002assembly}. Given the identical square shape of the pieces, their visual content is the only cue available for the reconstruction. Thus, as opposed to Freeman's and Garder's~\cite{freeman1964apictorial} work, most of the computer vision literature preferred to neglect the geometrical aspect of puzzle solving, although virtually all real-world puzzle-solving challenges involve fragments with different shapes, thus requireing both geometric and pictorial content as reconstruction cues.

While square jigsaw puzzles dominate the literature, studies on other types of puzzles, and in particular puzzles of arbitrary pieces, do exist. 
For example, Le et al.~\cite{le2019jigsawnet} presented a deep-learning puzzle solver where formally no explicit restriction is introduced on fragment shapes. In practice, however, their CNN detector was trained and tested only on perturbed rectangles, and geometry proved rich enough to effectively marginalize the contribution of the pictorial data.
Derech et al.~\cite{derech2021solving} explored archeological puzzle solving with piece shapes borrowed from dry mud patterns, though no attempt was made to analyze the possible shapes and their influence on the puzzle-solving process, and pairwise alignment was based on matching exitings and projected pictorial content. 
More formal and rigorous treatment was perhaps the Crossing-Cuts (CC) puzzles due to Harel et al.~\cite{harel2021crossing,harel2024pictorial}, where the puzzles are formed by cuts throughout a global polygonal shape, followed by an erosion process applied to each fragment. Excluding accidental cases, no 3 cuts are expected to meet at the same intersection point, and this geometric observation inspired a CC puzzle solver that searches for ``loops'' of precisely four pieces around joint vertices, followed by seeking ``loops of loops'' in a hierarchical fashion to construct large yet reliable puzzle assemblies~\cite{harel2024pictorial}. With a formal generation process, CC puzzles provide a unique opportunity to analyze the properties of non-trivial puzzles, but they are still shy of puzzles with pieces of general shapes encountered in real life.

\begin{figure}[h!]
    \centering
    \begin{tabular}{c|c}
        \includegraphics[width=0.25\textwidth]{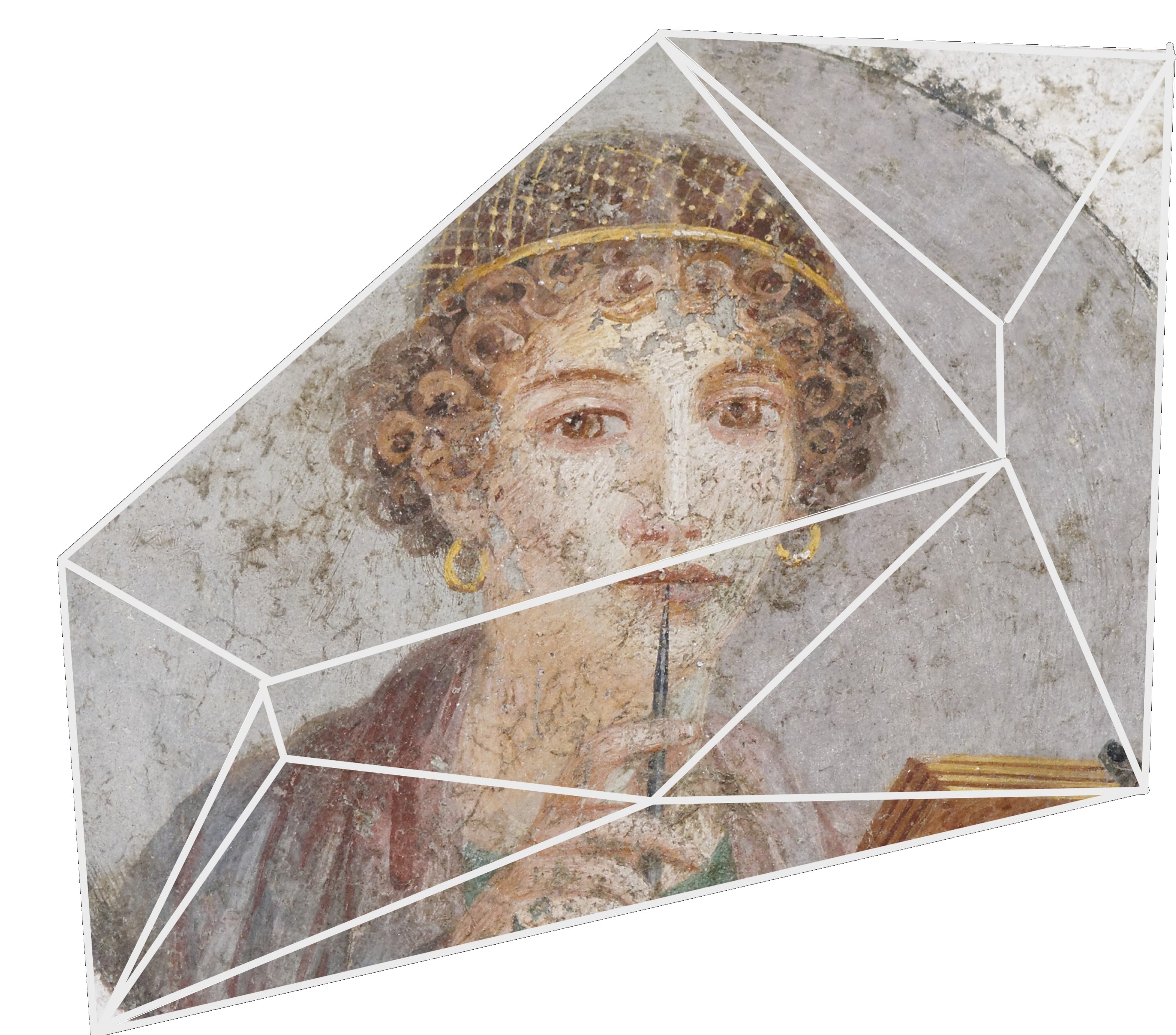} & 
        \includegraphics[width=0.30\textwidth]{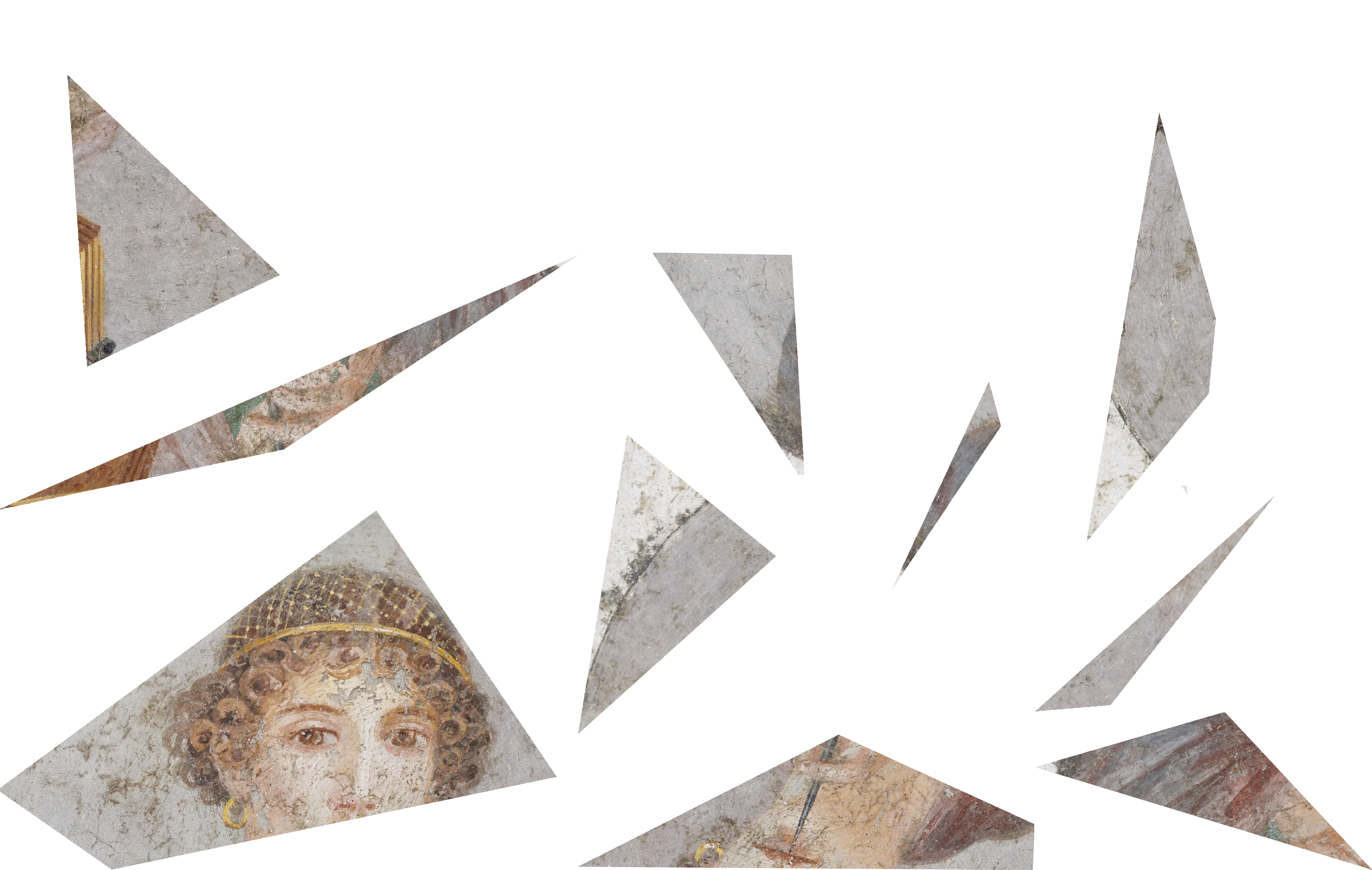} \\
        A & B
    \end{tabular}
    \caption{
    Convex partitions and puzzles.
    {\bf (A)} A Convex Partition graph superimposed on an image of a Pompeii fresco (free photo from Wikimedia, by Marie-Lan Nguyen). In such depictions, it is easy to see how each edge of each face/piece matches at most one edge of a different face/piece.
    {\bf (B)} A Convex Partition \textit{puzzle} created from the convex partition by randomly shuffling and transforming the pieces. The challenge of Convex Partition puzzle solvers is to reconstruct the coherent structure and image from A.}
    \label{fig:convex_partition_example}
\end{figure}

In this work, we take one step further and present a puzzle solver designed for a broader class of puzzles based on what is known as Convex Partitions~\cite{demaine2020computing}, as exemplified in Fig.~\ref{fig:convex_partition_example}. 
A convex partition for a given set of points in the plane (a.k.a the \textit{seed set}) is a planar graph that induces a convex polygonal subdivision of the points' convex hull. 
This subdivision must satisfy specific properties, and in particular, the graph must include all edges of the convex hull and its straight-line edges must form disjoint convex polygons whose interiors are empty (i.e., do not contain any points from the seed set). This formulation does not permit all conceivable puzzles, but it is a strict generalization of CC, the resultant puzzles are very easy to construct, and they are much closer to real-world applications. Solving them is thus an important stepping stone to practical general puzzle solvers.

\section{Related Work}

Puzzle-solving methods can be classified using different criteria. This first is their type, either \textit{aprictorial puzzles}, which rely solely on the geometric shape of the pieces, or \textit{pictorial puzzles}, which also (and sometimes solely) leverage visual content for reconstruction. Here we will focus on the latter type which is also more relevant to the computational vision community. 

A second classification criterion for puzzle solving is the puzzle's geometric structure, that also reflects its generation process. 
In  this realm, \textit{square jigsaw puzzles}
are by far the most popular setup in the computational literature~\cite{toyama2002assembly,fei2007image,zhao2007puzzle,murakami2008assembly,alajlan2009solving,cho2010probabilistic,pomeranz2011fully,yang2011particle,andalo2012solving,sholomon2013genetic,adluru2015sequential,gallagher2012jigsaw,mondal2013robust,son2016solving,son2018solving,paikin2015solving} and as mentioned above, given the identical shape of the pieces their algorithmic solution rely exclusively on visual content, as geometry provides no useful information. 
Square puzzle solvers typically compute compatibility scores between pieces and use greedy or heuristic methods to maximize global consistency~\cite{cho2010probabilistic,pomeranz2011fully}. variations include handling unknown piece orientations~\cite{gallagher2012jigsaw,mondal2013robust,sholomon2014generalized,son2014solving,yu2015solving,son2016solving,son2018solving,rika2019novel}, missing pieces~\cite{gallagher2012jigsaw,mondal2013robust,paikin2015solving,son2016solving,son2018solving}, and noisy data~\cite{mondal2013robust,son2014solving,yu2015solving,son2018solving,rika2019novel,brandao2016hot}, while performance peaks at puzzles with tens of thousands of pieces (e.g.,~\cite{sholomon2014generalized}). Search and optimization techniques for solutions range from minimal spanning trees~\cite{gallagher2012jigsaw}, to quadratic programming~\cite{andalo2012solving}, genetic algorithms~\cite{toyama2002assembly,sholomon2013genetic}, relaxation labeling~\cite{khoroshiltseva2021jigsaw,vardi2023multi}, and deep learning approaches~\cite{rika2019novel,scarpellini2024diffassemble,liu2024solving,giuliari2024positional}, to name but a few.

\textit{Unrestricted puzzles} 
lack formal constraints on piece shapes, allowing arbitrary geometry and piece adjacencies. Used briefly during the inception of the field in the literature ~\cite{freeman1964apictorial,kong2001solving,zhou2025pairingnet, jin2023interactively} they typically use curve-matching techniques and greedy algorithms, and typically experimented on small-scale problems only. The few recent attempts such as JigsawNet~\cite{le2019jigsawnet} incorporate pictorial data and neural networks to handle larger puzzles with up to 400 pieces. However, even these approaches often rely on constrained train and test data, alluding more to perturbed rectangles rather than general piece shapes. Similar apictorial challenges have also been explored in archaeological puzzle-solving contexts~\cite{zheng2024reunion,ostertag2020matching,papaodysseus2002contour,sizikova2017wall,andalo2016automatic}. 

Although not fully unrestricted, \textit{(partially) restricted puzzle} models define non-trivial formal generation processes that significantly generalize square puzzles while allowing a wider range of applications with some ability for rigorous analysis. Examples include the brick wall puzzles~\cite{gur2017square} where pieces are rectangles of different dimensions, and CC puzzles~\cite{harel2021crossing,harel2024pictorial} of polygonal pieces.
In this work, we propose and address a much more general class of partially restricted puzzles that brings us even closer to most conceivable real-life applications of puzzle solving.

\section{Convex partition puzzles - Formulation}
\label{section:formulation}

A Convex Partition puzzle \textit{problem} can be conceptualized as obtaining a random shuffle of the faces of a Convex Partition graph and seeking to reconstruct the original (unknown) structure. Thus, the vertices of all pieces originate from the seed set points, and the edges of pieces are derived from the graph's edges.

With this in mind, let $P=\{p_1,p_2,...,p_N\}$ be the set of a Convex Partition puzzle pieces, where the sequence of vertices of each piece $p_i$ is $(v_i^1,v_i^2,...,v_i^{N_i})$, such that $v_i^1,v_i^2,...,v_i^{N_i}\in\mathbb{R}^2$ and $N_i$ is the number of vertices of $p_i$. We assume w.l.o.g. that this sequence of vertices is sorted counter-clockwise around the center of mass of the piece. The sequence of $p_i$ edges is thus
{\scriptsize
\begin{equation*}
(e_i^1,...,e_i^{N_{i-1}},e_i^{N_i}) \defeq 
((v_i^1,v_i^2),...,(v_i^{N_{i-1}},v_i^{N_i}),(v_i^{N_i},v_i^1))\;.
\tag{1}
\end{equation*}
}

A puzzle \textit{solution} is defined primarily by a set of 
rigid transformations $\{(R_1,t_1),...,(R_N,t_N)\}$ of all pieces to their reconstructed pose, where  $R_i\in\mathbb{R}^{2\times2}$ is the orthonormal rotation matrix of piece $p_i$ and $t_i\in\mathbb{R}^2$ is its corresponding translation vector. Each transformation $(R_i,t_i)$ maps the vertices of $p_i$ to their final positions in the reconstructed puzzle, i.e., to $(R_iv_i^1+t_i,R_iv_i^2+t_i,...,R_iv_i^{N_i}+t_i)$.

While the transformations are often enough to reconstruct the solution in a visual sense, it is often constructive to also explicitly recover the pieces' neighborhood relationships, namely a graph representation that represents the direct (candidate or true) neighboring relationships between pairs of different pieces in the reconstructed puzzle, or what is knows as the~\textit{mating graph}~\cite{freeman1964apictorial,harel2024pictorial} (see below).

\section{Noisy Puzzles}
\label{section:noisy_puzzles}

The abstract Convex Partition puzzle problem assumes (unrealistically) perfect geometric information, and in particular the location of piece vertices and thus the length of piece edges. Hence, assuming a convex partition of a random seed set, the probability of more than two edges in the puzzle having identical lengths is nil (cf. Fig.~\ref{fig:convex_partition_example}). Under such conditions, the reconstruction of the puzzle becomes trivial, and can be obtained by simply matching pieces with edges of identical length. 

Practical real-life puzzles, however, are never that sterile and geometric information is typically noisy. While geometric noise can be modeled in various ways, here we adopt a popular choice in the prior art to represent material \textit{erosion}.
Specifically, inspired by recent proposals~\cite{harel2024pictorial}, each vertex $v_i^k$ of piece $p_i$ is pushed inside \textit{into} the piece by $\epsilon_i^k$ distance to obtain the noisy vertices $(v_i^1+\epsilon_i^1,v_i^2+\epsilon_i^2,...,v_i^{N_i}+\epsilon_i^{N_i})$ where $||\epsilon_i^k|| \sim U(0,\varepsilon)\ $ and $\varepsilon$ is bounded by some percentage $\xi$ of the puzzle diameter $D$ (distance between furthest vertices), namely, $\varepsilon=\xi \cdot D$. As analyzed by Harel et al.~\cite{harel2024pictorial}, the maximum length difference between matched edges can be $\pm 4\varepsilon$. 
We later use this bound in our algorithm to identify the set of potential matches for each edge. Since the cardinality of this set is typically large, and all but one potential match are false positives,  solving and reconstructing noisy puzzles is a nontrivial task. Indeed, a naive complete solver will need to check all candidate matches for each edge, clearly an untractable (exponential) search process.

\section{Convex Partition Puzzles Solver}
\label{sec:puzzle_solver}

To cope with the combinatorial complexity of noisy Convex Partition puzzle problems, we propose a greedy solver based on the following general steps.
First, the solver formulates coarse geometric pair compatibilities by comparing the lengths of all piece edges and keeping in the mating graph only those within the $\pm 4\varepsilon$ difference. 
Next, it relaxes the initial compatibilities by incorporating pictorial affinity and retaining the most promising matings. 
Then, the solver optimizes the final mating selection and computes the Euclidean transformations of the solution with a global optimization that mimics the minimization of a physical spring-mass energy similar to Harel et al.~\cite{harel2024pictorial}. The details of all these steps are described in the following subsections.

\subsection{The Matings Graph}
\label{section:matings_graph}

The mating graph is a graph whose nodes correspond to the edges of the pieces, while its edges represent immediate neighboring relationships. To avoid terminological confusion between edges of pieces and edges of the graph we refer to later as \textit{links}. The nodes store the length of their corresponding piece edges, which will become handy during reconstruction.

Importantly, we define two types of links: \textit{mating links} and \textit{piece links}. Mating links connect nodes from \textit{different} pieces, and $M$ will denote the set of all those links in the graph. On the other hand, piece links define neighboring relationships \textit{within} a piece as follows from its polygonal representation. In other words, two nodes will be linked by a piece link if and only if these nodes represent two adjacent edges of the same puzzle piece. This subset of links can be pre-computed just once ahead of time by making a single pass on the graph representation of all pieces (hence a linear $\BigO(n)$ operation, where $n=\sum_i{N_i}$). 

Since the reconstructed puzzle is a Convex Partition, every mating link in the ground truth mating graph links exactly \textit{two} edges of two different faces, and thus every edge of a puzzle piece should match at most \textit{one} other edge of \textit{identical length} from another piece (cf. Fig.~\ref{fig:convex_partition_example}). We call this the \textit{monogamy mating constraint}. Excluded are edges that belong to the convex hull of the seed set and thus have no match. 

In addition to initializing the piece links as above, we also initialize a complete mating subgraph, i.e., we construct every mating link between each node of one piece to all other nodes of all other pieces. This initialization represents the initial uncertainty of which piece edges are matched in the final solution.

\begin{figure}[h!]
    \centering
    \begin{tabular}{c c}
         \includegraphics[width=0.35\textwidth]{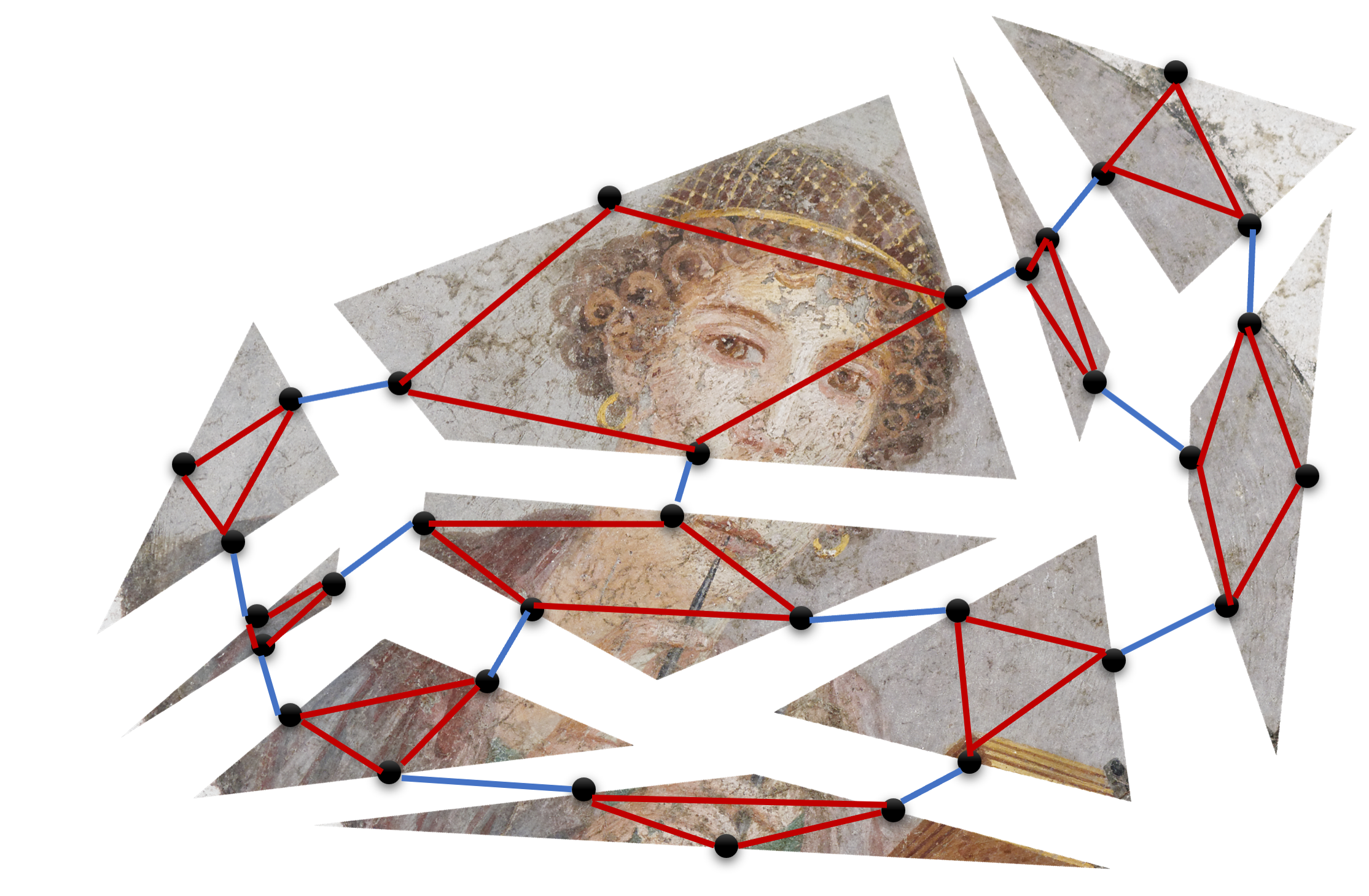} &
         \includegraphics[width=0.35\textwidth]{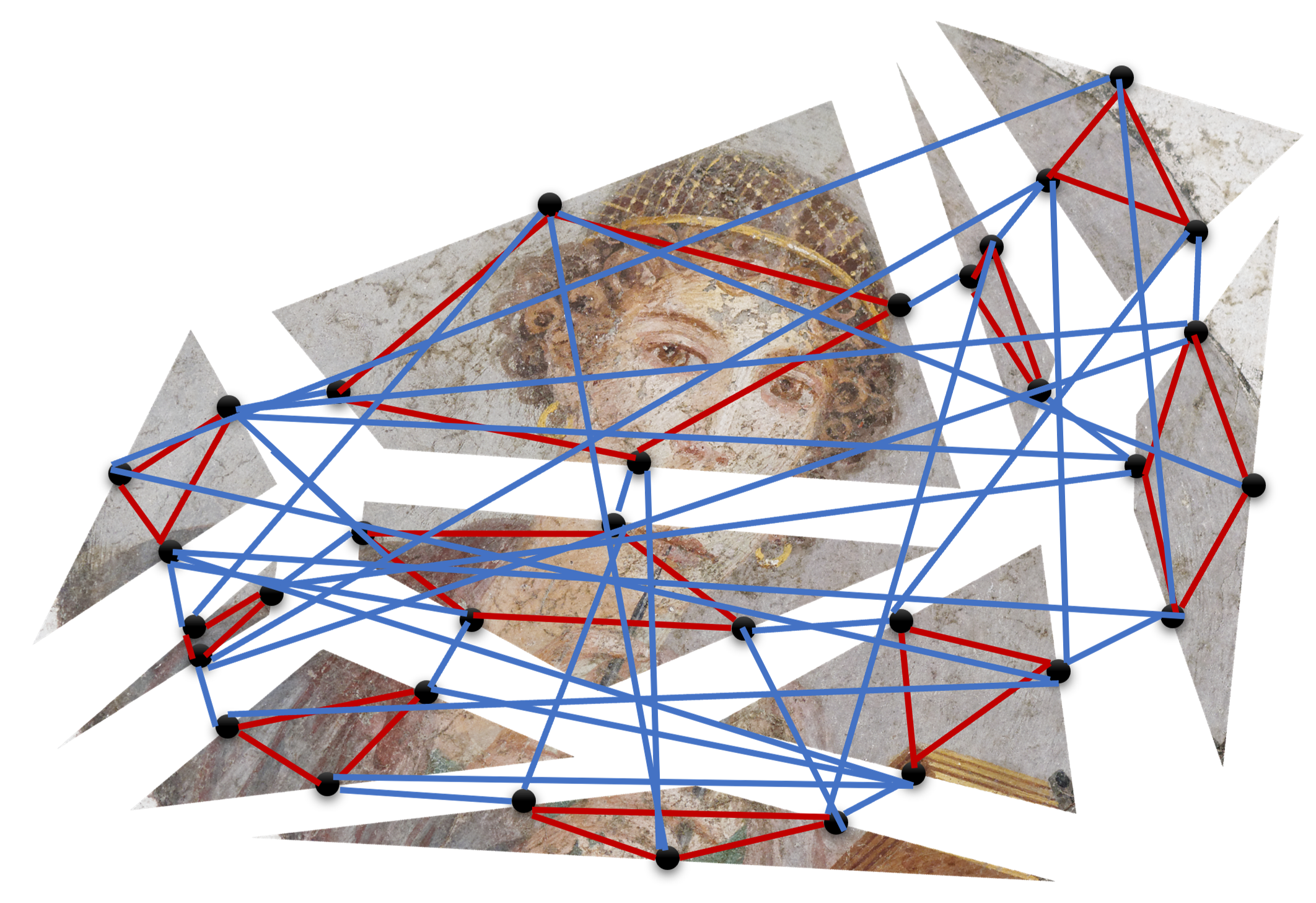} \\ 
         A & B
    \end{tabular}
    \caption{The mating graph of the convex partition from Fig.~\ref{fig:convex_partition_example}. The nodes (i.e., the edges of the pieces) are depicted as black points. Piece and mating links are colored red and blue, respectively.
    {\bf (A):} A noiseless nonaccidental puzzle, where edges of pieces have at most one mating link.
    {\bf (B): } A noisy puzzle, were nodes might have many mating links. The full mating graph for this noisy puzzle contains many more mating links than illustrated, but to avoid clutter we show only a subset.}
    \label{fig:mating_graph}
\end{figure}

If the puzzle is noiseless and nonaccidental (i.e., at most 2 edges have identical lengths), the complete subset of mating links can be immediately diluted based on the monogamy mating constraints such that each node will have at most one mating link -- the one that links it to the only other node with identical length. The result is exemplified in Fig.~\ref{fig:mating_graph}A. Note that nodes representing boundary edges have only piece links and are always of degree 2. 

If the puzzle is noisy, we can again filter out some mating links, though now we must retain all those links whose nodes are within the $\pm 4\varepsilon$ length difference. Unfortunately, this often preserves several mating links for many nodes, as exemplified in Fig.~\ref{fig:mating_graph}B.
Note that a similar situation might occur with noiseless but accident puzzles.

\subsection{Pictorial Compatibility}
\label{sec:pictorial_compatibility}

We develop a pictorial compatibility and compute it for each mating link (that survived the geometric filtering) and documented in the graph as its weight. 
Applying this compatibility will later help to refine the coarse geometric compatibility from section~\ref{section:matings_graph} and to further reduce false positive mating links.

\begin{figure}[h]
    \centering
    \begin{tabular}{cc}
    \includegraphics[width=0.24\textwidth,height=0.347\textwidth]{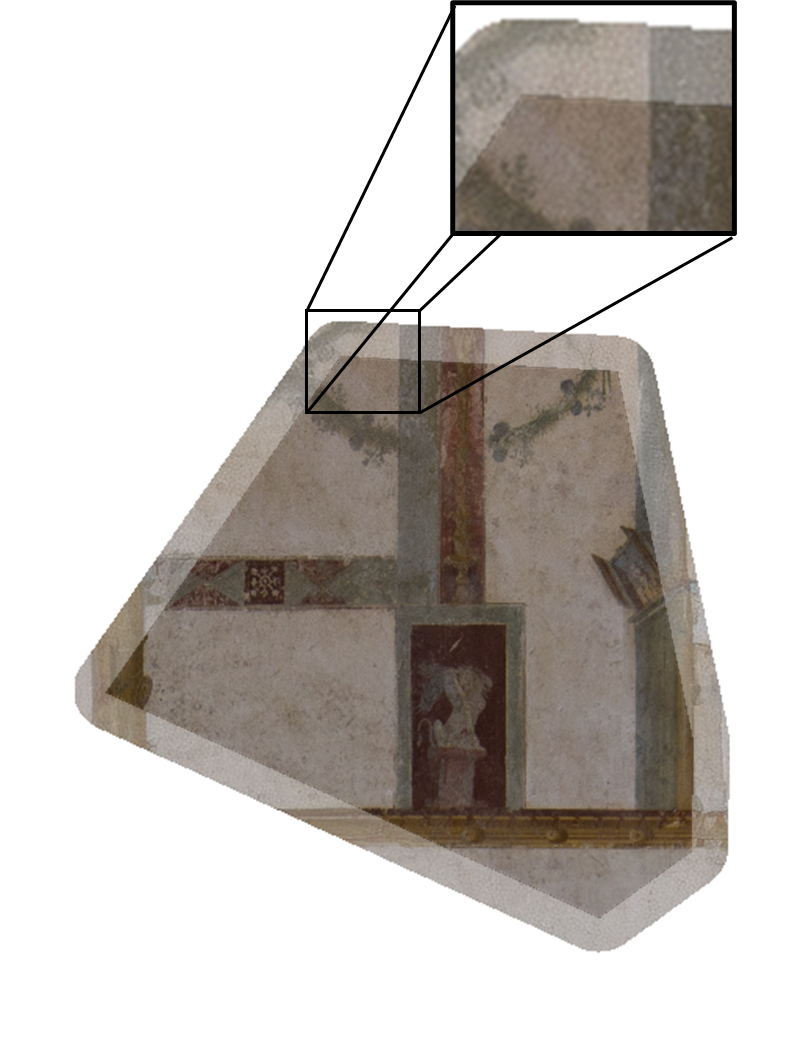} &
        \includegraphics[height=0.34\textwidth]{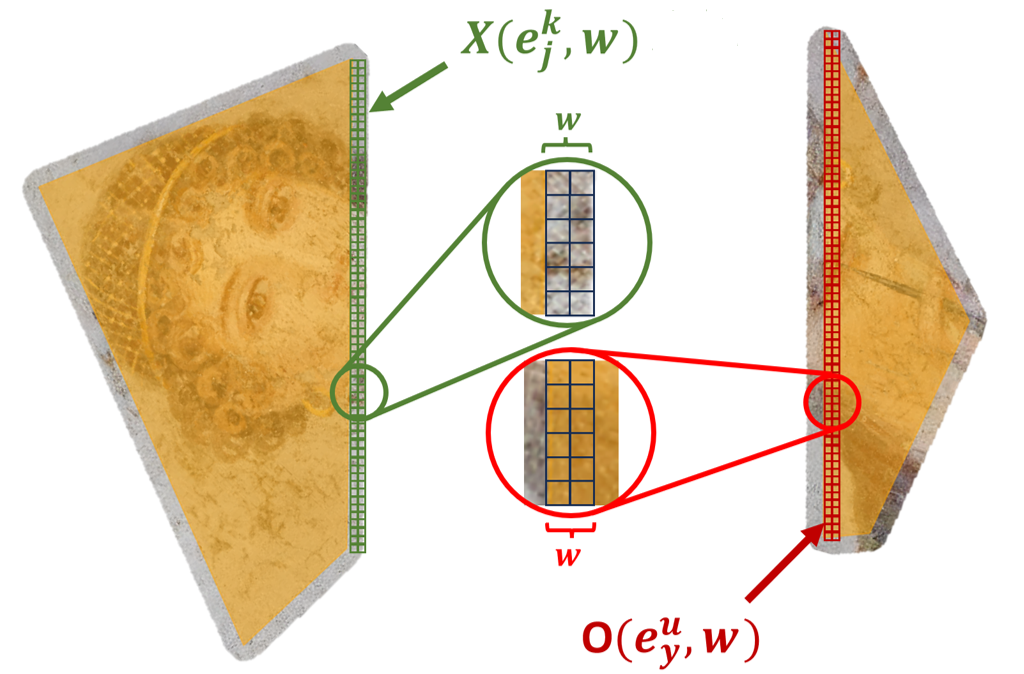}\\
        A & B
    \end{tabular}
    \caption{Pictorial extrapolation and compatibility. 
    {\bf (A)} Extrapolation of pictorial data into the eroded zone (and beyond) using Stable Diffusion~\cite{Rombach_2022_CVPR}. The extrapolated band is shown in lighted shades for clarity).
    Fragment taken from a recent fresco fragment dataset~\cite{elkin2025recognizing}. 
    {\bf (B)} The compatibility of a mating link, in this case between the two vertical edges, is computed between $X(e_k^j,w)$ and $O(e_y^u,w)$, depicted in green and red, respectively. Note that the former is extrapolated data while the latter is genuine. For clarify the sketch uses $w=2$ though in practice it was larger. The colored (orange) region demarcates the genuine data of the eroded piece while the extrapolated band is in grayscale. Here, $l(e_y^u)$  and $l(e_j^k)$ are roughly equal.
    }
    \label{fig:pixel_grids}
\end{figure}

Clearly, the computation of pictorial compatibilities needs to cope with missing pictorial content (due to the erosion) exactly where it is needed the most, i.e., in regions where pieces are supposed to abut and pictorial information presumably continues across piece boundaries. Of course, even if pieces were not eroded and no information was missing, some scheme of matching pictorial content from different pieces is still needed.
To handle the missing pictorial information due to erosion, we thus extrapolate the exiting pictorial content around a piece to a band of predefined width using the Stable Diffusion model by Rombach et al.~\cite{Rombach_2022_CVPR}. Fig.~\ref{fig:pixel_grids} exemplifies one such extrapolation.

With all pieces extrapolated, the solver now compares the extrapolated content around one edge to the genuine pictorial content of the other edge across the corresponding mating link. This comparison is made by ``overlapping`` the images and evaluating pixel-wise similarities.

More formally, let $X(e_k^j,w)$ represent the pixel grid of the extrapolated content for edge $e_k^j$ (of piece $p_k$),  after both were rotated to align with the pixel grid, as in Fig.~\ref{fig:pixel_grids}). $w$ is the grid’s width (i.e., its ``thickness''). Let $l(e_k^j)$ denote $e_k^j$'s length.
Similarly, let $O(e_y^u,w)$ denote the pixels grid of the \textit{original} content of edge $e_k^j$ where the width of the grid is again $w$ (and its length along the edge $e_y^u$ is $l(e_y^u)$). Unlike $X(e_k^j,w)$, $O(e_y^u,w)$ is extends \textit{inside} the piece.

As a preprocessing step, the pixel grids (both extrapolated and original) are normalized channel-wise (i.e., separately for R,G,B) by their means. %
Then, to score the compatibility between a mating link between edges $e_k^j$ and $e_y^u$, we essentially score the similarity of $X(e_k^j,w)$ and $O(e_y^u,w)$. 
However, since the edges' lengths are not usually equal, we use a ``sliding window'' mechanism as a workaround, with which the shorter grid slides over the longer one, and a measure of compatibility is averaged over all relative positions.
For that, assume w.l.o.g that $\Delta=l(e_k^j)-l(e_y^u)>0$, so $X(e_y^u,w)$ is smaller and slides over $O(e_k^j,w)$ (otherwise, the same applies by switching rules). 
Moreover, let $W_{\delta}$ be a sub-window of size $l(e_y^u)$ at offset $\delta$ from the beginning of $O(e_k^j,w)$.
With this, the similarity $F$ between $O(e_k^j,w)$ and $X(e_y^u,w)$ is defined as:
{ \scriptsize
\begin{equation*}
    F \left( O(e_k^j,w),X(e_y^u,w) \right) = 
    \dfrac{\sum_{\delta=0}^{\Delta} \left( X(e_y^u,w) \circ W_{\delta} \right)}{\Delta+1}
    \tag{2}\label{equation:sliding_window}
\end{equation*}
}

The operation $\circ$ between pixel grids $M_X$ and $M_O$ is simply
{\scriptsize
\begin{equation*}
    M_X \circ M_O = \frac{\mathbf{1}^T(M_X \odot M_o)\mathbf
    {1}}{||M_X|| \cdot ||M_o||}
    \tag{3}\label{equation:pictorial_dot_product}
\end{equation*}
}
where $\odot$ is the element-wise product (a.k.a Hadamard product~\cite{horn2012matrix} or Schur product~\cite{davis1962norm}),$\mathbf{1}$ is an all 1 column vector and $||\cdot||$ is the Forbenius norm. The dot product $M_X \circ M_O$ is thus the cross-correlation of the pictorial signals normalized to their mean.

With the above basic operations, the pictorial compatibility of a candidate mating $(e_k^j,e_y^u)$, is defined as
{
\scriptsize
\begin{equation*} 
    \begin{aligned}[t]
        C(e_k^j, e_y^u) = & \; \frac{1}{2} \cdot F \Big( 
            O(e_k^j, w),
            X(e_y^u, w) 
        \Big) + \\
        & \frac{1}{2} \cdot F \Big( 
            O(e_y^u, w), 
            X(e_k^j, w) 
        \Big)
    \end{aligned}
    \tag{4}\label{equation:pictorial_comp}
\end{equation*}
}
namely, it is the average of scoring the extrapolated data of $e_y^u$ against the original data $e_k^j$ and the scoring of the extrapolated data of  $e_k^j$ against the original data $e_y^u$). Averaging these two scores compensates for imperfections in the extrapolation mechanism, especially when it was done 
much better for one piece compared to the other.


Finally, all compatibilities are normalized to $[0,1]$ (based on their minimal and maximal values throughout).
Mating links with compatibility below a selected empirical threshold are then filtered out from the graph.

\subsection{Global Spatial Optimization}
\label{section:spatial_optimization}

\begin{figure}[h!]
    \centering
    \begin{tabular}{ccc}
        \includegraphics[width=0.22\textwidth]{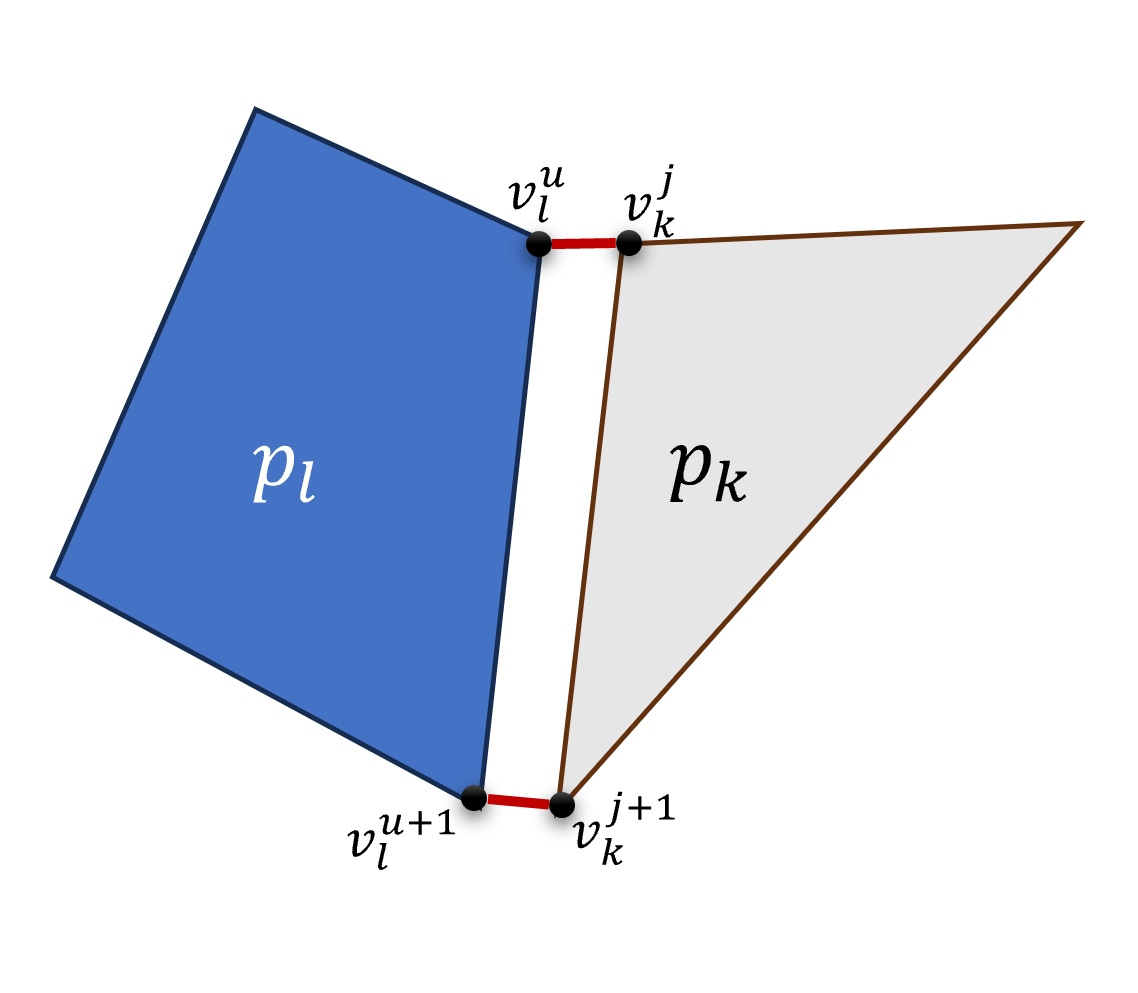} &  \includegraphics[width=0.22\textwidth]{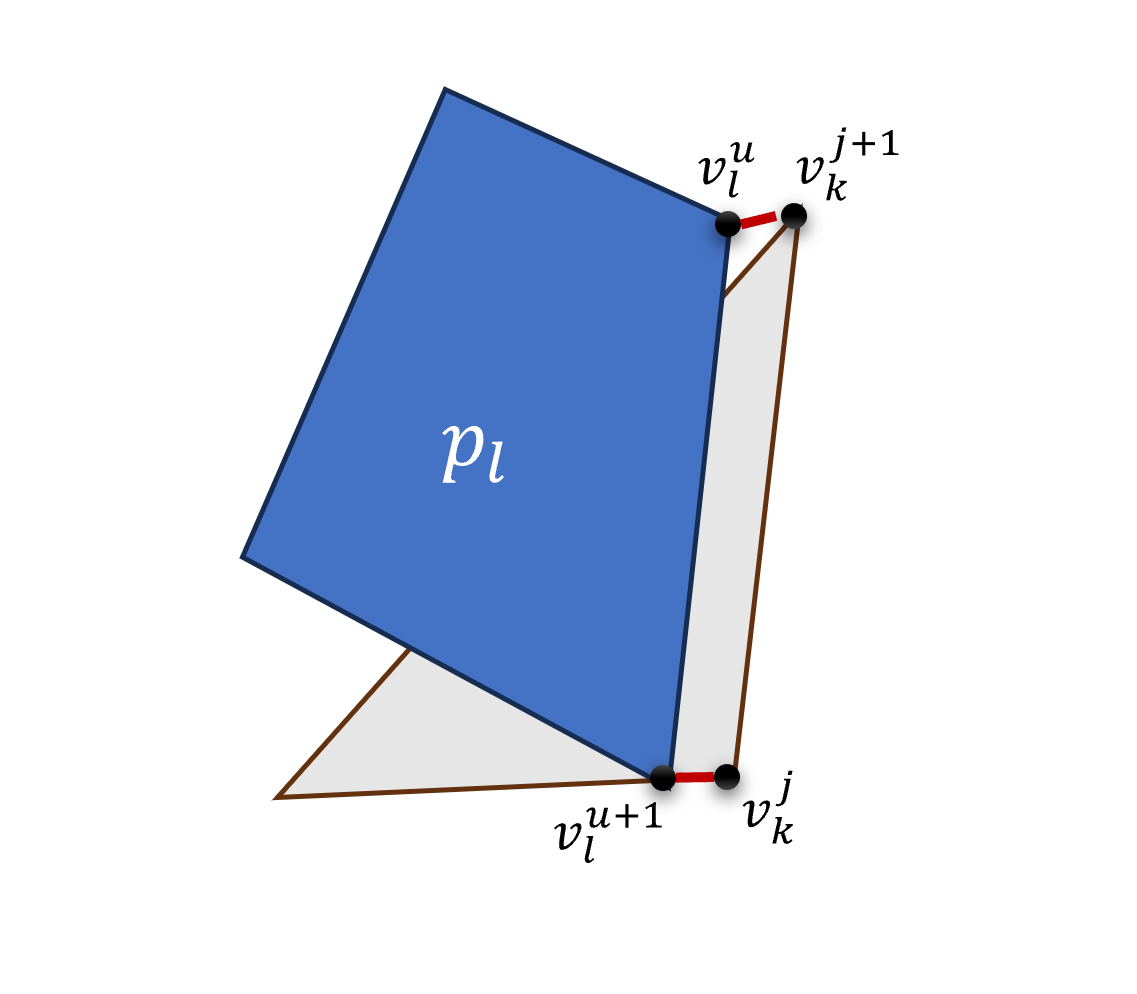} &
        \includegraphics[width=0.22\textwidth]{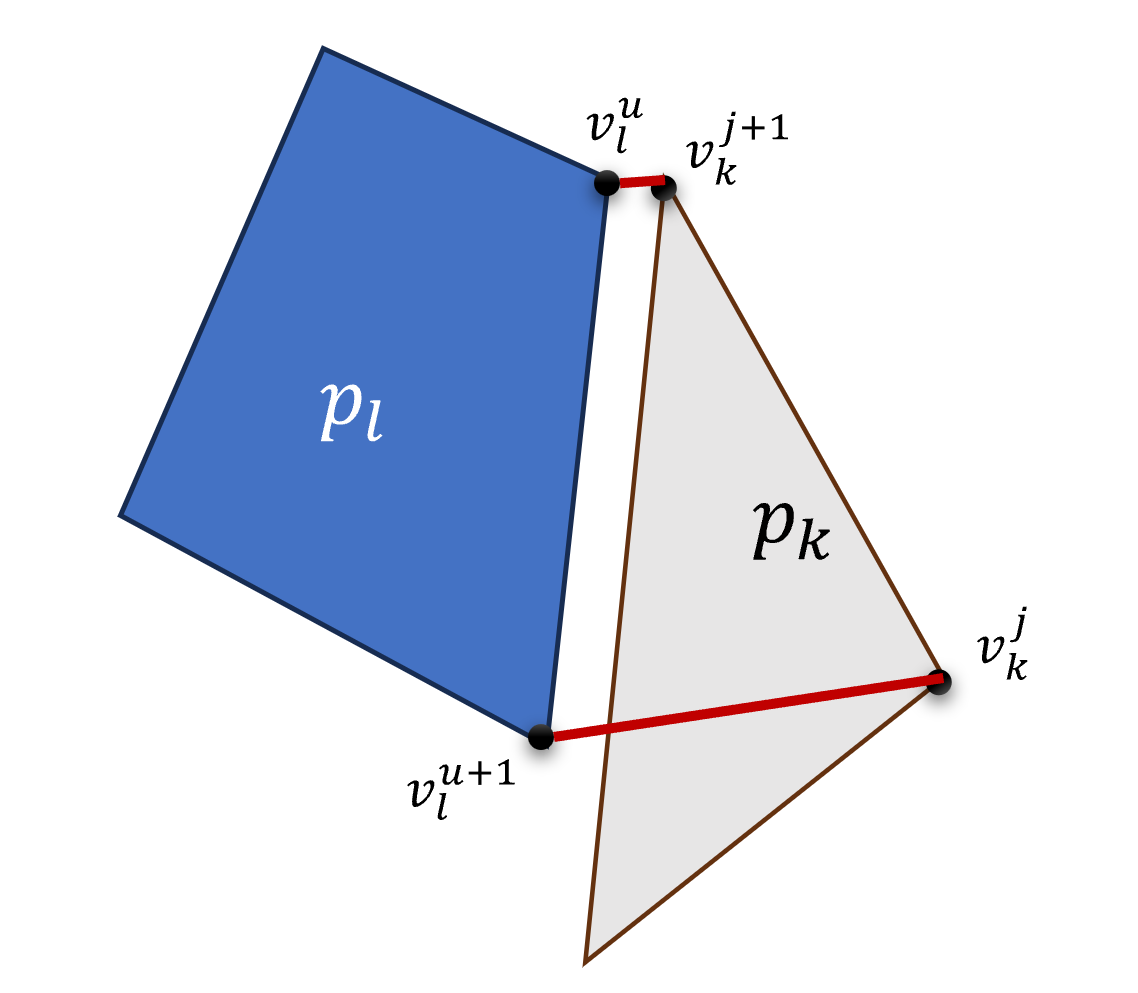} \\
        A & B & C
    \end{tabular}
    \caption{Illustration of matching vertices of the mating edges $(v_k^j,v_{k+1}^j)$ and $(v_l^u,v_{l+1}^u)$ of the pieces $p_l$ (blue) and $p_k$ (gray), respectively. The corresponding distances/springs between vertices) are colored red.
    {\bf (A):} The (correct) mating vertices $v_k^j$, $v_l^u$ and $v_{k+1}^j$,$v_{l+1}^u$. 
    {\bf (B):} Incorrectly matching $v_{k+1}^j$ to $v_l^u$ and $v_k^j$ to $v_{l+1}^u$  entails overlap between the pieces when the distance between vertices (or equally, the energy of the connecting springs) is minimized.
    {\bf (C):} Incorrect matching is possible without overlaps only if the combined distances (i.e., spring lengths) is non negligible.
    }
    \label{fig:mating_verties}
\end{figure}

While the previous steps aimed at generating a mating graph with the most promising mating links (both geometrically and pictorially), the formulation of a puzzle solution must also include the Euclidean transformation of the pieces. However, when this is required for multiple noisy eroded pieces, it is not simply a matter of aligning piece edges and rather requires non-trivial optimization. 
Let the vertices of the edges in a mating link  $\left(e_k^j,e_l^u \right)$ be $(v_k^j,v_k^{j+1})$ and $(v_l^u,v_l^{u+1})$, respectively. 
As Fig.~\ref{fig:mating_verties} suggests, there are only two configurations for pairing $e_k^j$ and $e_l^u$. Either $v_k^j$ is paired with $v_l^u$ and $v_k^{j+1}$ is paired with $v_l^{u+1}$, or $v_k^j$ is paired with $v_l^{u+1}$ and $v_k^{j+1}$ is paired with $v_{l}^u$. Clearly, only one configuration is desired (as in Fig.~\ref{fig:mating_verties}A) because the other one entails an overlap between the pieces (Fig.~\ref{fig:mating_verties}B) unless the matched edges are kept excessively apart (Fig.~\ref{fig:mating_verties}C). We call the matched vertices in the desired configuration of a mating (i.e., the one that lacks overlapping)  as its \textit{mating vertices}.

Given a set of pieces $P=\{p_1,...,p_N\}$, a good solution is a set of transformations $T=\{(R_1,t_1),...,(R_N,t_N)\}$ that simultaneously minimize the Euclidean distance for all mating vertices $(v_k^j,v_l^u)$ in $M$~\cite{harel2024pictorial}. Formally, we seek: 
{\scriptsize
\begin{equation*}
    \label{equation:objective_of_springs}
    argmin_{\; T} \ \frac{1}{2}\sum_{(v_k^j,v_l^u) \in M} {|| (R_kv_k^j + t_k) - (R_lv_l^u + t_l)||}^2
    \tag{5}
\end{equation*}
}
where $(R_k,t_k) \in T$ is the transformation for piece $p_k$ and $(R_l,t_l) \in T$ is the transformation for the piece $p_l$. 
Importantly, the vertices of the same piece must have the exact transformation, and this is not a simple least squares problem.  Inspired by Harel et al.~\cite{harel2024pictorial}, we thus consider a numerical process that minimizes this expression by abstracting it as a multi-body spring-mass system whose equilibrium state satisfies a similar expression.
For that we consider every puzzle piece to be a 2D rigid body with uniform density and total mass proportional to their area, and every pair of mating vertices of a mating under consideration is connected by a spring having zero length and constant elasticity (i.e., all the springs have identical spring constants). The potential energy of this spring-mass system is expressed as  $\sum_l \frac{1}{2} k x_l^2$ ,
where $x_l$ represents the extension or compression of spring $l$ from its natural length, and $k$ is the spring constant. When $k=1$, this expression is identical to Eq.~\ref{equation:objective_of_springs}, and it is often handled numerically in the physics community with proper simulation (for which, in our case, we utilized the open-source physics engine Box2D~\cite{Box2D}). 

In practice, such a numerical solution process begins by assigning random transformations (translations and orientations) to all pieces in the arena, except for one anchor piece that will resolve the ambiguity of the solution up to a rigid motion (i.e., a global Euclidean transformation). Hence, we select the piece with the most number of mating links and assign it an infinite mass so it does not move and remains anchored to the center of the arena.  While this anchoring does not affect the generality of the solution, it does speed up the convergence of the numerical simulation. 

\begin{figure}[h]
    \centering
    \begin{tabular}{ccc}
        \includegraphics[width=0.30\textwidth]{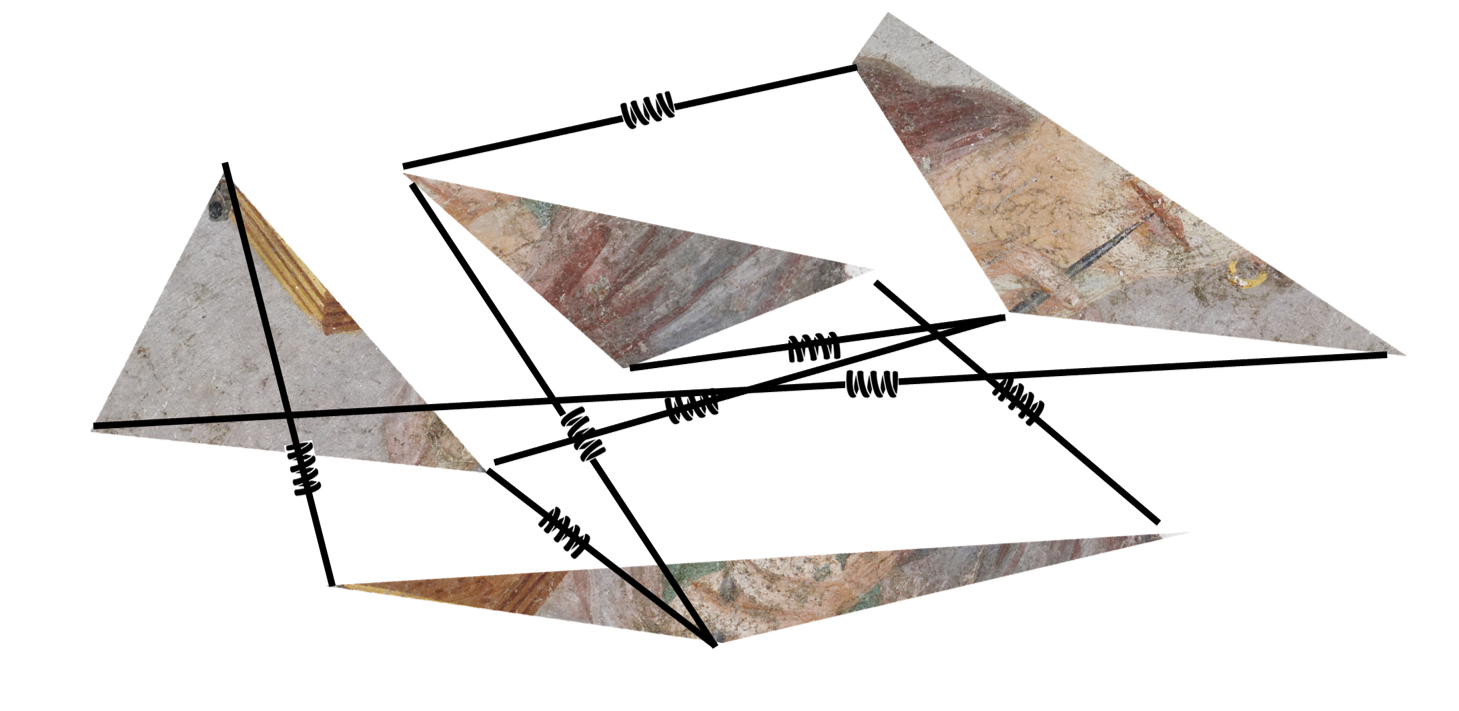} & \includegraphics[width=0.30\textwidth]{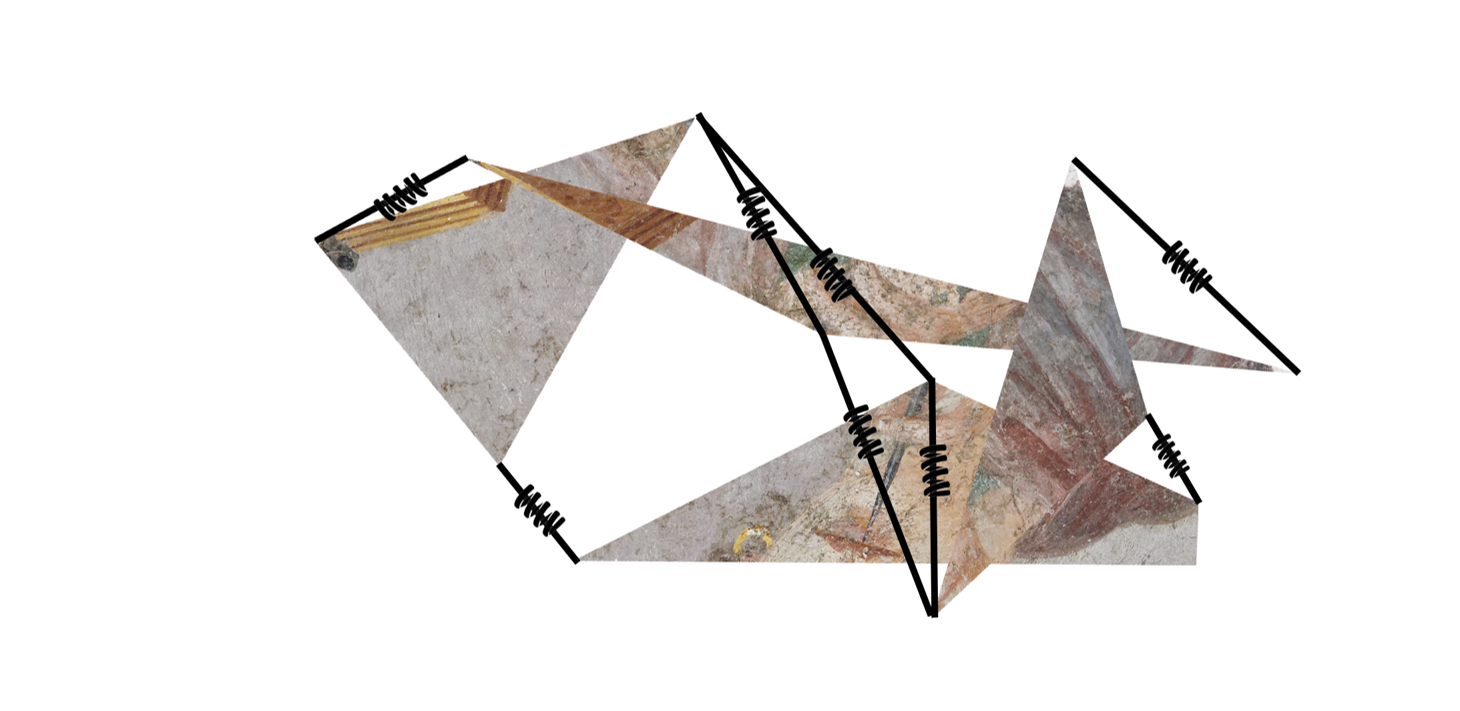}  & \includegraphics[width=0.30\textwidth]{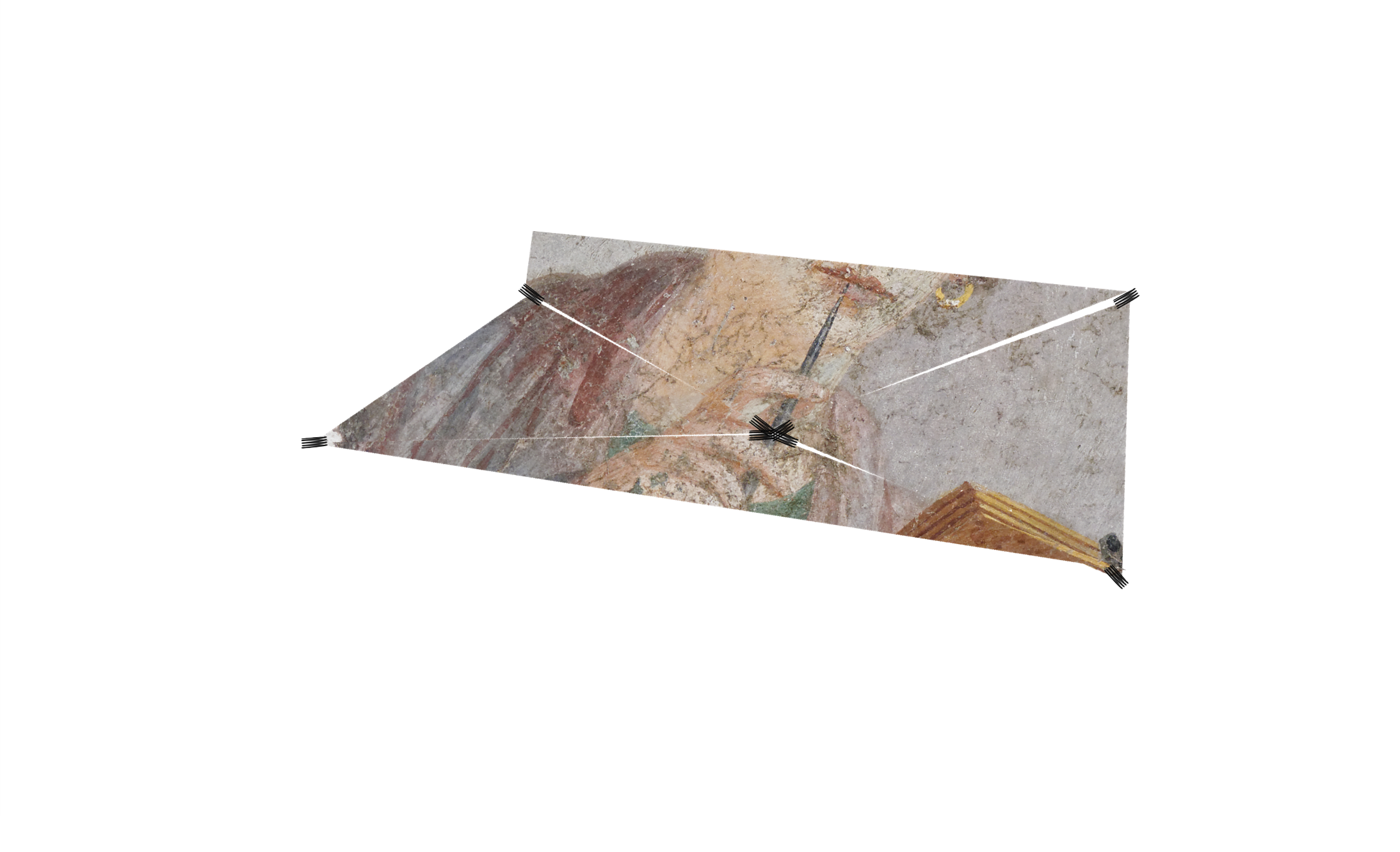}  \\
        A & B & C
    \end{tabular}
    \caption{Illustration for the springs-mass system simulation on four pieces . 
    {\bf (A):} The initial state of the pieces and a pair of springs for each mating link that is considered. 
    {\bf (B):} An intermediate state of the pieces as they get closer to their final pose while artificially allowed to overlap each other.
    {\bf (C):} The final state of the bodies after convergence and while overlapping is no longer allowed.
    }
    \label{fig:springs_demonstration}
\end{figure}

After initialization. the physical system is allowed to evolve based on the force inflicted dynamically by the springs, and with some energy loss due to friction (i.e., damping) it converges to its minimal energetic state. The process is composed of two phases. First, the pieces are allowed to overlap during motion (i.e., physical collisions are disabled) to allow the simulation to overcome positional deadlocks due to the initial random configurations. This converges to an approximate solution where some overlaps are still observed. Then a second phase is initiated, this time after disabling the possibility of overlaps between the pieces, and the simulation proceeds until a final \textit{physically plausible} convergence is obtained. 
%
Fig.~\ref{fig:springs_demonstration} illustrates selected frames (i.e., time steps) for a small assembly of four noisy pieces at the beginning, somewhere in the middle, and as soon as the system converges to its equilibrium state  .

\subsection{Global Mating Optimization}

The spatial optimization above requires a mating graph to operate on. To solve the puzzle, our solver thus hypothesizes such a graph and uses it to compute the transformations. This process involves iteratively refining the initial mating graph (from Sec.~\ref{sec:puzzle_solver}) by removing links until it represents a hypothesized unambiguous solution. 
we construct this solution by building small local compact assemblies and progressively merging them into a unified whole, arguably the most important part of our proposed solver.

\begin{figure}[h!]
    \centering
    \begin{tabular}{c c c}
        \includegraphics[width=0.28\textwidth]{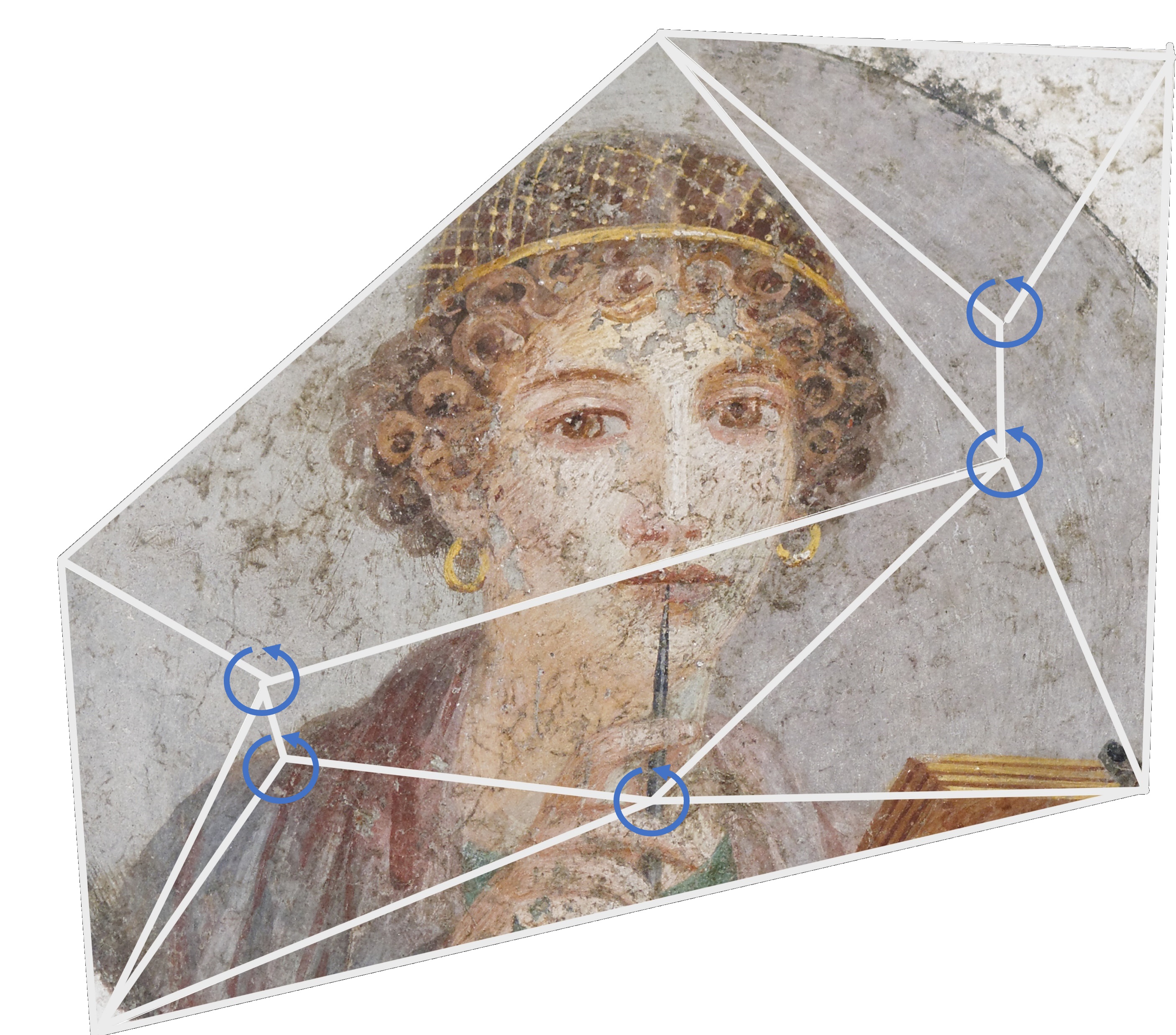} & \includegraphics[width=0.28\textwidth]{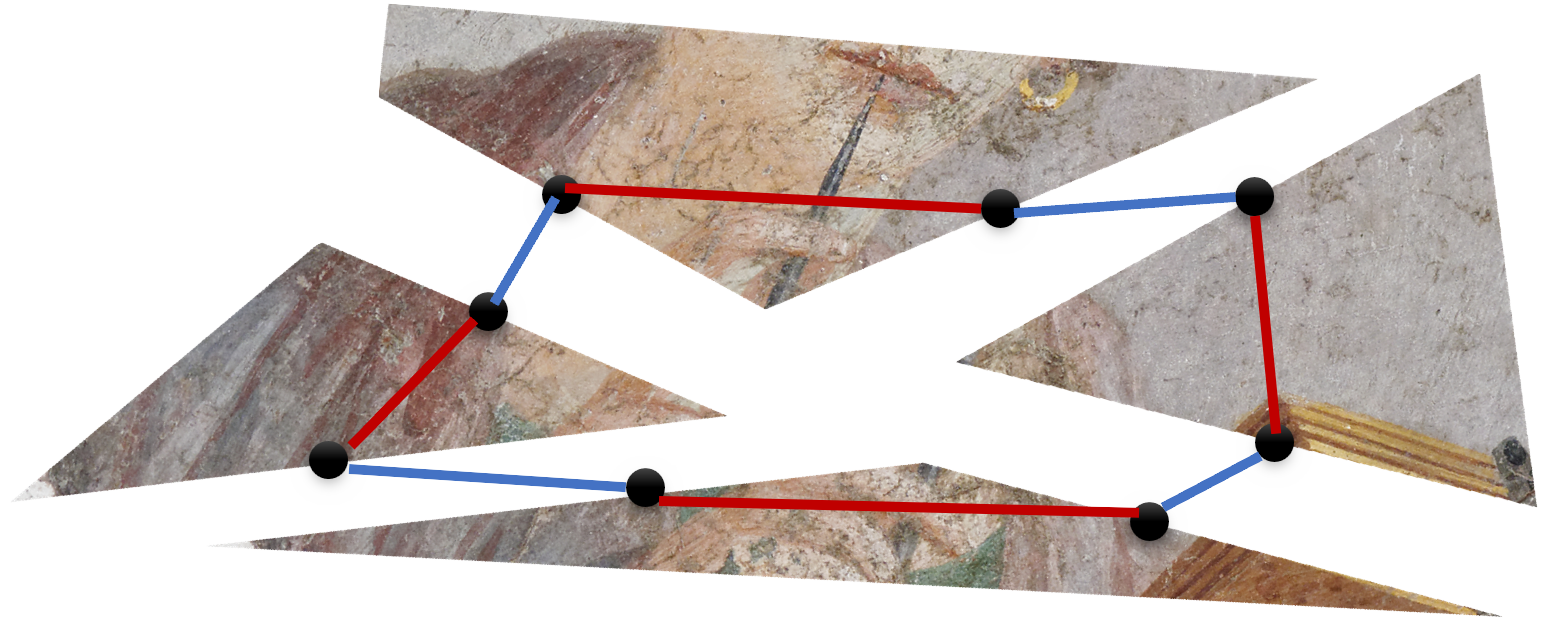} & \includegraphics[width=0.30\textwidth]{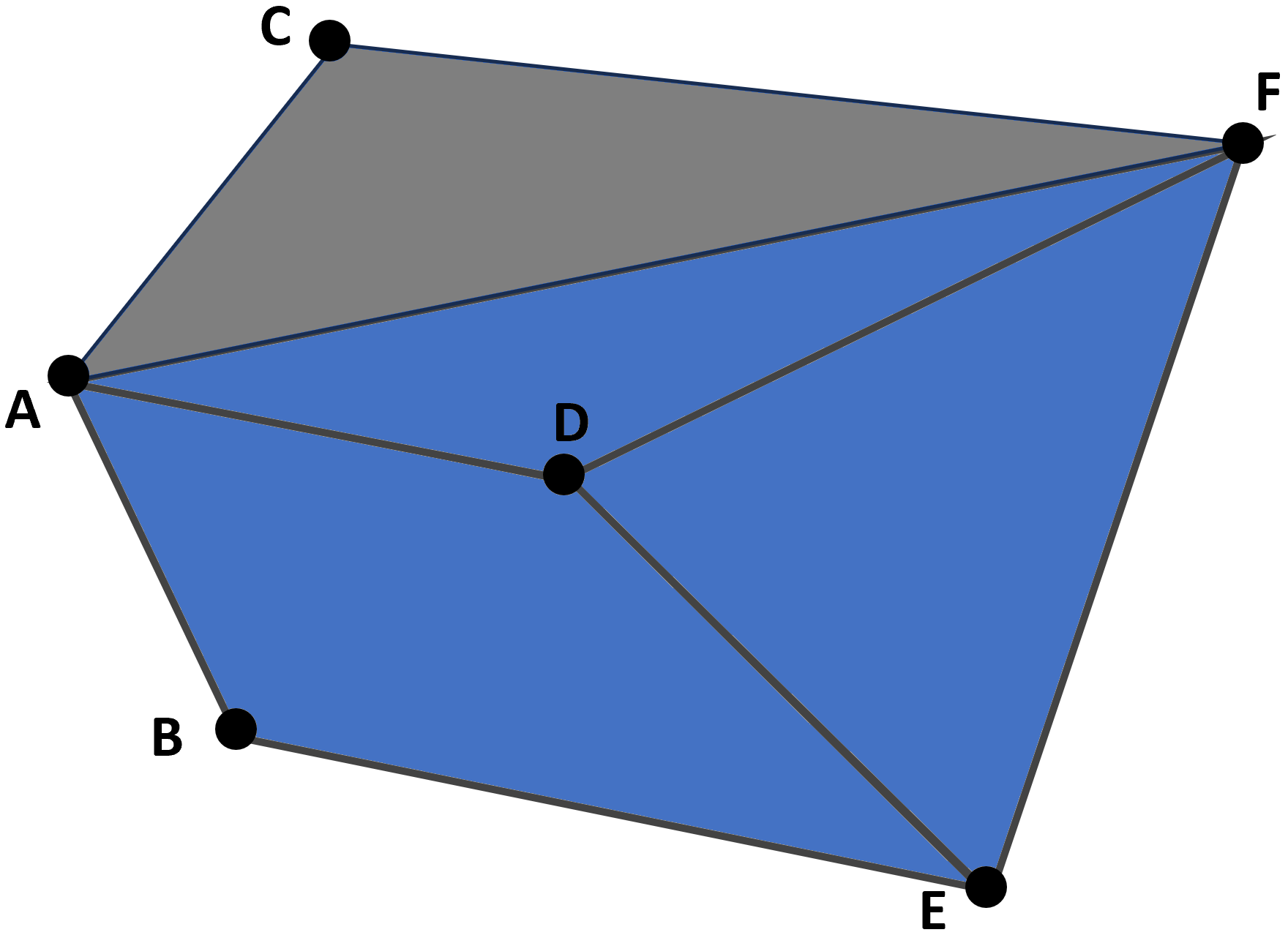}\\
        A & B & C
    \end{tabular}
    \caption{Cycles in mating graphs.
             {\bf (A):} Circular arrows show cycles (of different size) of faces around the interior vertices. Searching for such cycles and reconstructing them locally is the first step in solving the puzzle.
             {\bf (B):} The graph representation of the lower cycle in panel~A. Note the alternation between pieces links (red) and mating links (blue). It is this type of cycles that the solver looks for in the mating graph. {\bf (C):} Boundary pieces with no internal (i.e., seed set) vertices are sentenced to participate in no cycle whatsoever. Here, no vertex of $\triangle ACF$ is internal (say, like point $D$), and thus it cannot participate in any cycle.}
    \label{fig:loops}
\end{figure}

The computation of these local assemblies is based on an inherent property of the Convex Partition graphs by which each \textit{internal} vertex of the graph (i.e., a point inside the convex hull of the seed set) is where three or more faces of the graph meet (see Fig.~\ref{fig:loops}A). Thus, one should expect to find \textit{cycles} in the mating graph where mating links and piece links \textit{alternate} (see Fig.~\ref{fig:loops}B). 
This is true for any partition that can be described as a graph, but unlike in square puzzles (e.g.,~\cite{son2018solving}), or even CC puzzles~\cite{harel2021crossing,harel2024pictorial}, in Convex Partition graphs constraints are available neither on the number of pieces nor on the interaction of piece edges at those internal vertices, so a different scheme must be used. 

Since each cycle should revolve around a common vertex in the convex partition, it can be detected by starting from one node in the mating graph and pursuing a consistent counter-clockwise path for 3 or more alterations (and thus an even number) of nodes and edges before closing a cycle~\cite{cormen2022introduction}. While mating links can lead anywhere, piece links are much more constrained and are, in fact, selected uniquely. 
Indeed, once we traverse a mating link, the subsequent piece link is selected such that it leads to a counter-clockwise edge in the piece (Recall from Sec.~\ref{section:formulation} that edges are sorted this way in our piece representation, so no outstanding computation is needed).

Successful cycles require consistency between several mating links of different pieces, which significantly reduces the likelihood of incorrect combinations. However, false positives can still occur because the geometric compatibilities are tested in the presence of noise, and pictorial compatibilities are not foolproof. To further reduce false positives, our solver scores each cycle by attempting to reconstruct its spatial structure using the spatial optimization as described in Sec.~\ref{section:spatial_optimization}. 
The scoring itself is split into two components, each emerging from the different phases of the spatial reconstruction, the first while piece overlapping is allowed and the second when it is prohibited.
Technically, given set $P$ of pieces in a cycle and an operator \texttt{$A(\cdot)$} that returns the area of a given piece, the score of the first phase is defined as a Dice coefficient~\cite{dice1945measures} that sums the relative overlap area of each piece \texttt{$p_i$} to all other pieces in the cycle. Formally:
{\scriptsize
\begin{equation*}
    \tag{6}\label{equation:enabled_overlapping}
    Q_{ol}(P)=\sum_{p_i \in P}\frac{|A(p_i) \cap \left( \cup_{p_j \neq p_i}{A(p_j)} \right)| }{|A(p_i)|}
\end{equation*}
}
The score of the second phase, when overlaps are prohibited, quantifies the distances between pieces once the process converges to its minimal energy:  
{\scriptsize
\begin{equation*}
    \tag{7}\label{equation:disabled_overlapping}
    Q_{td}(M)=\sum_{(v_k^j,v_l^u) \in M}{||v_k^j - v_l^u)||}^2
\end{equation*}
}
where $(v_k^j,v_l^u)$ are the mating vertices in \texttt{$M$}.

With this in mind, the total score of a cycle is obtained by the weighted sum of the two measures above, namely
{\scriptsize
\begin{equation*}
    \tag{8}\label{equation:loop_penalty}
    Q_{cycle}(P,M) = \alpha \cdot Q_{ol}(P) + (1- \alpha) \cdot Q_{td}(M)
\end{equation*}
}
The left component in Eq.~\ref{equation:loop_penalty} penalizes cycles for greater overlapping regions between their pieces since correct assemblies are expected to lack it. The right expression, on the other hand, expresses the objective function of the mating vertices from Eq.~\ref{equation:objective_of_springs}.

Once the cycles are ranked and sorted by their $Q_{cycle}$ score (where lower is better), the solver identifies the first cycle in which the absolute difference between its score and the score of its successor cycle in the ranking is higher than a predefined threshold $\tau$. The successor cycle and the rest of the following cycles are removed from the list. Then, mating links that appear in no cycle in the pruned list are removed from the mating graph altogether. 

Unlike the CC~\cite{harel2021crossing,harel2024pictorial} subspace of puzzles, a piece in a Convex Partition puzzle may not be able to take part in \textit{any} cycle. As Fig.~\ref{fig:loops}C illustrates, this can happen even under sterile conditions, and in particular for boundary pieces without internal vertices. We denote those pieces as \textit{individual pieces}, and place them at the end of the ranking.

Once the list of cycles (and individual pieces as mentioned above, if present) is finalized, the solver turns to combine them into larger assemblies, referred to as \textit{aggregates}. An item in that list serves as a basic aggregate, and when an aggregate is integrated into a larger aggregate, they both lose their individual identity and can no longer combine separately into a different aggregate. 

The integration of two aggregates into a larger one is done by taking the union of their node and link sets, and is allowed if 
(i) at least one of the aggregates includes a piece that the other does not,
(ii) the aggregates have mutual pieces, and
(iii) the aggregates don't conflict with assigning a mating link to a node and thus don't break the monogamy mating constraint. 
In practice, these conditions amount to 
(i) checking that their node sets are different, 
(ii) verifying that the intersection of their node sets is non-empty, and 
(iii) no node exceeds degree 3 (and specifically, no more than one mating link).

We note that the search for cycle combinations is not random but rather determined by the ranking of the cycles by $Q_{cycle}$, with attempts involving higher-ranked cycles preceding lower-ranked ones. Technically, this is done by maintaining a sorted queue $Q$ of unaggregated cycles that are initialized by all cycles (and individual pieces, if present) found earlier.
The aggregates are then formed by iterating over the sorted list, removing the highest ranked cycles, and scanning the rest of the list from top to bottom to find other cycles that can be merged to it by the criteria above. A cycle that cannot be merged is skipped but keeps its position in the list. A cycle that can be merged is removed from the list and merged into the aggregate, and since this opens up a new opportunity for merging skipped cycles, the scanning is restarted from the top of the list again. In the worst case, when each scan always merges the lowest ranking cycle, an iteration is quadratic in the number of cycles. Once it is over, the aggregate found is kept aside, and a new iteration begins from the now-new top ranked cycle in the $Q$. Note that if an iteration yields no merging, the top ranking cycle is removed from the list anyway, and declared an aggregate of its own. Since the process shrinks $Q$ after every iteration, it is guaranteed to terminate as soon as $Q$ is empty.

Once $Q$ is empty, we are left with a collection of aggregates. While the mating links of each aggregate represent unique matches between the nodes they connect, different aggregates may suggest different (and thus conflicting) mating links for the same nodes in the mating graph. To obtain the final mating graph that satisfies the monogamy constraint, we thus turn back to the original mating graph obtained in Sec.~\ref{sec:pictorial_compatibility} and use the aggregates to filter out all but one mating link in each node. Aggregates that were formed earlier are prioritized (since they were formed by higher-ranked cycles) and their mating links are retained in the final graph first. Note that individual pieces (cf. Fig.~\ref{fig:loops}C again) do not break the scheme since their own mating links are already in the graph and will not be filtered out.

Once all conflicts are resolved, the mating graph is finalized and is used to define the set of constraints (i.e., springs) for the spatial optimization of the full puzzle, and thus to obtain the final solution.

\section{Dataset}

Since no prior work on Convex Partitions (or for that matter, any general polygonal) puzzles exists, we have created a benchmark dataset to test our proposed solver on. The dataset contains three versions of 25 puzzles (i.e., in total 75 puzzles): A noiseless version, a version where $\xi=0.1\%$, and a version where $\xi=0.25\%$. Recall that this parameter sets the noise relative to the diameter of the entire puzzle so even small values can significantly affect individual pieces. The size of the puzzles ranges from 6 pieces to 40 pieces, where the median is 16 and the average is 17.9. 

All images used to construct our dataset were retrieved from the Wikimedia Commons repository, specifically from the “Category:Frescoes” collection, using the MediaWiki API. Each puzzle contains the images and polygonal representations of its pieces and the ground-truth transformations and matings.
Needless to say, many more puzzles, with any number of seed points or other parameters, can be created easily on demand, and our synthesis code is shared for the benefit of the community.

\section{Experimental Evaluation}

Recall from Sec.~\ref{sec:puzzle_solver} that a puzzle solution includes the mating graph of the reconstructed puzzle and the Euclidean transformation of each piece. Following the proposed evaluation measures from the literature~\cite{harel2024pictorial}, we assess performance using two main metrics.
The mating graph of the proposed solution is evaluated against the mating graph of the ground truth using recall, precision, and $F1$ on the set of mating links, while the Euclidean transformations of pieces are evaluated by the normalized overlap area between pieces in the proposed solution vs. their ground truth counterparts~\cite{harel2024pictorial}, i.e.,
{\scriptsize
\begin{equation*}
        Q_{pos}(R_1,t_1,\dots,R_n,t_n) = \sum_{i=1}^n{w_i \frac{|A(\overline{p_i}) \cap A(R_i p_i + t_i)|}{|A(p_i)|}} 
    \tag{9}\label{equation:qpos}
\end{equation*}
}
where \texttt{$(R_i,t_i)$} is the Euclidean transformation suggested by the solver for piece $p_i$ and the weight $w_i = |A(p_i)| / \sum_{k=1}^{n} |A(p_k)|$
is proportional to the piece area, emphasizing the greater importance of larger pieces in the overall puzzle shape. $\overline{p_i}$ is $p_i$ after being transformed according to the ground truth.

\begin{table}[b!]
    \centering
    \scalebox{0.85}{
    \begin{tabular}{||c c c c c||}
        \hline
        Noise level $\xi$ & Precision $\uparrow$ & Recall $\uparrow$ & F1 $\uparrow$ & \texttt{$Q_{pos}$} $\uparrow$\\
        \hline \hline
        \texttt{$0.0\%$} & 100\% & 74.86\% & 84.95\% & 37.7\% \\
        \hline
        \texttt{$0.1\%$} & 86.38\% & 75.44\% & 79.92\%  & 38.13\% \\
        \hline
        \texttt{$0.25\%$} & 78.11\% & 62.73\% & 69.23\% &  30.6\% \\
        \hline
    \end{tabular}}
    \caption{The solver's performance on the proposed dataset with different levels of geometric noise. Note how the noise level gracefully affects performance, as expected.}
    \label{table:solver_results}

\end{table}

\begin{table}[b!]
    \centering
    \scalebox{0.85}{
    \begin{tabular}{||c c c c||}
        \hline
        Solver &  Precision $\uparrow$ & Recall $\uparrow$  & \texttt{$Q_{pos}$} $\uparrow$\\
        \hline \hline
        Ours & 61.21\% & 77.42\% &  14\% \\
        \hline
        Harel et al.~\cite{harel2024pictorial} & $73.41$\% & $86.94$\% &  $52.35$\% \\
        \hline
    \end{tabular}}
    \caption{Our solver's results on the dataset of CC pictorial puzzles published by Harel et al.~\cite{harel2024pictorial}.}
    \label{table:competition}

\end{table}

\begin{figure}[b!]
    \centering
    \begin{tabular}{c c c}
        \includegraphics[height=0.22\textwidth]{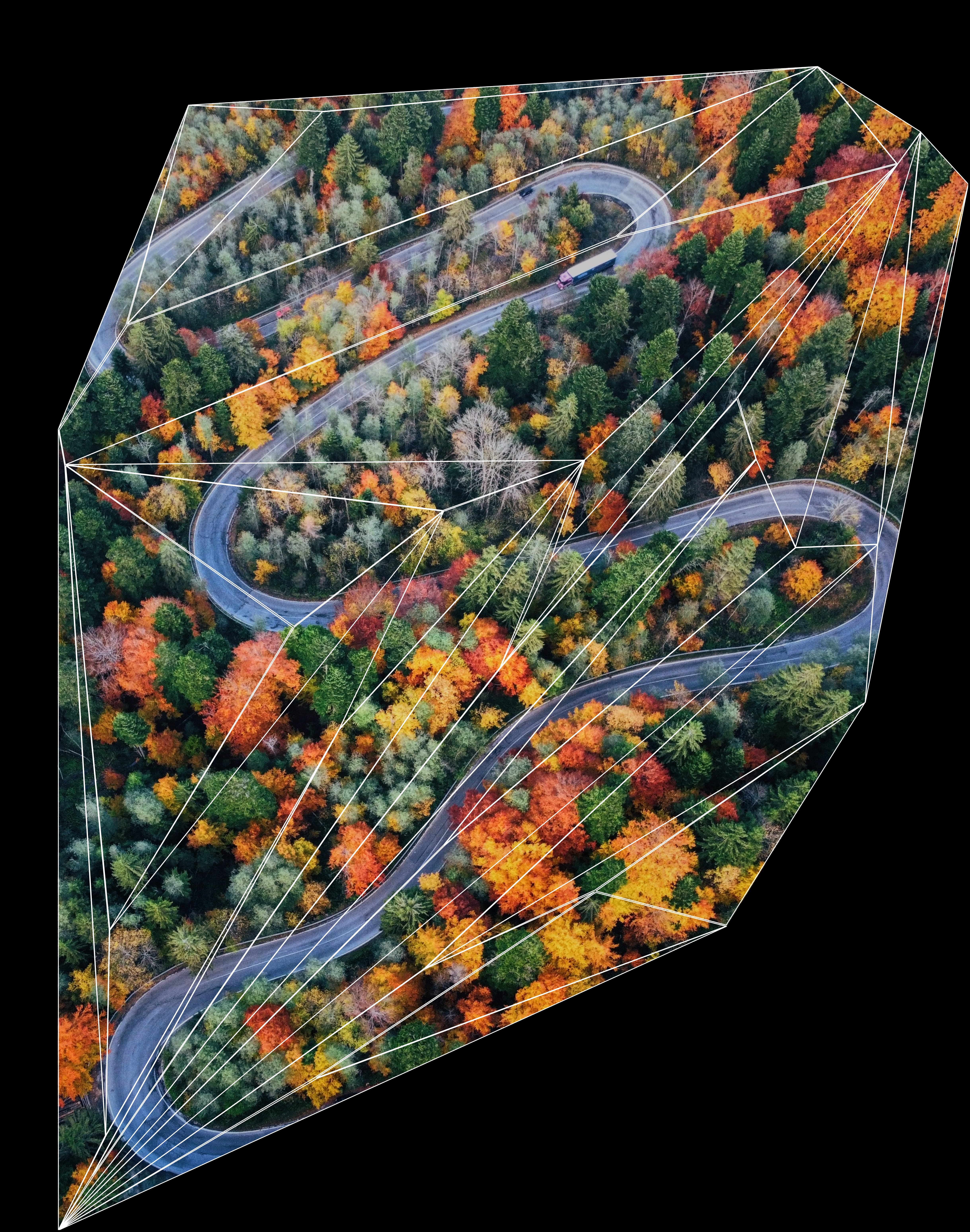} &         \includegraphics[height=0.22\textwidth]{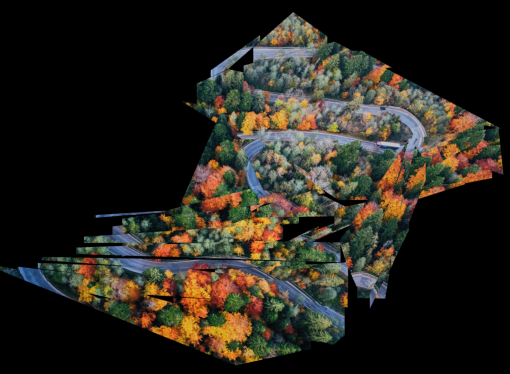} & \includegraphics[width=0.38\textwidth]{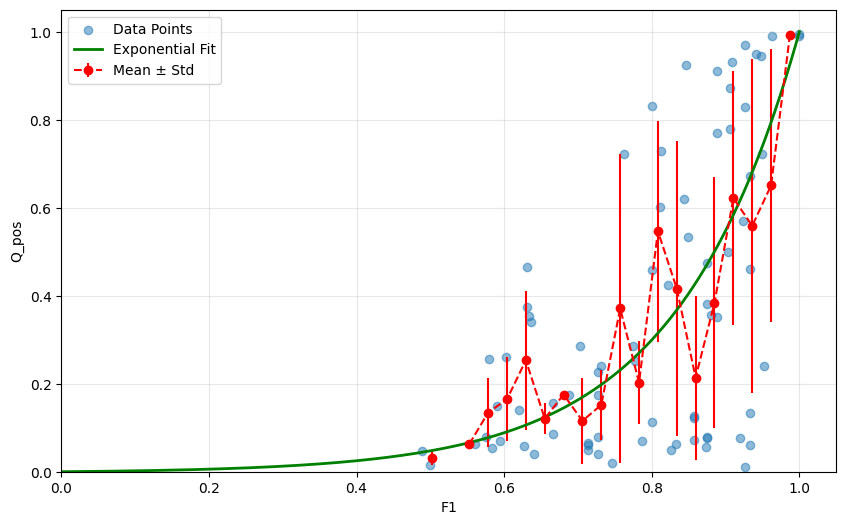} \\
        (A) & (B) & (C)
    \end{tabular}
    \caption{Demonstration of the conservative behavior of $Q_{pos}$. 
    {\bf (A)} Ground truth puzzle with $38$ pieces and noise level $\xi=0.1\%$.
    {\bf (B)} Reconstructed puzzle.  
    The solver reconstructed several large aggregates successfully but failed to merge them correctly. Visually, the result is decent, but while the precision and recall are $0.946$ and $0.898$, respectively, the $Q_{pos}$ score is $0.481$ as it tends to degrade rapidly even in the presence of a few mistakes that affect the coherency of the global reconstruction. This is also observed in Table~\ref{table:solver_results}. 
    {\bf (C):} Plotting $Q_{pos}$ against $F1$ demonstrates the conservative behavior of the former. Note the rapid degradation of $Q_{pos}$ as soon as $F1$ drops below the perfect score, and the convergence of the former to rather low values even for decent values of the latter.
    Blue points are actual values. Red points and bars are mean and STD values using 20 bins of width 0.025 in the range $[0.5,1]$. The fitted exponential model (in green curve) is $f(x)=\frac{e^{5.9987x}}{e^{5.9987} - 1}$.   
    }
    \label{fig:gloria_puzzle_qualitative}
\end{figure}

Table~\ref{table:solver_results} shows the performance measures when our proposed solver is applied to the dataset, and the parameters set to $w=5$, $\tau=50$, and $\alpha=0.5$. All experiments were conducted on a Windows 11 desktop equipped with a 12th Gen Intel Core i7-12700H processor (2.30 GHz, 14 cores, 20 threads), 16 GB of RAM, and a 64-bit architecture. Unfortunately, to our best knowledge, no other solver in the literature can handle convex partition (or general polygonal) puzzles, so comparison to a prior art is not yet possible.

Performance based on all measures is generally very good, and the precision and recall degrade gracefully with the level of geometric noise. 
$Q_{pos}$ appears to degrade more rapidly, but as shown in Fig.~\ref{fig:gloria_puzzle_qualitative}A,B, and argued already by its developers~\cite{harel2024pictorial}, this measure should be considered with caution for its conservative behavior.  
Fig.~\ref{fig:gloria_puzzle_qualitative}C plots $Q_{pos}$ against $F1$ on the proposed dataset,
to show that, except for occasional outliers, $Q_{pos}$ is generally monotonic with $F1$  but lags significantly behind it (before accelerating rapidly near the higher bound). Put differently, as soon as $F1$ falls below the perfect score, $Q_{pos}$ degrades fast prior to settling down, a behavior characteristic of this measure~\cite{harel2024pictorial}.

\begin{figure}[h!]
    \begin{tabular}{c|c|c|c}
        \setlength{\tabcolsep}{0pt}
        \includegraphics[height=0.18\textwidth]{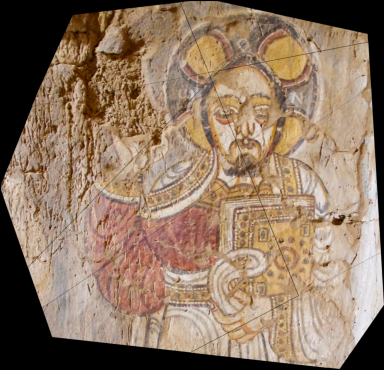} & \includegraphics[height=0.18\textwidth]{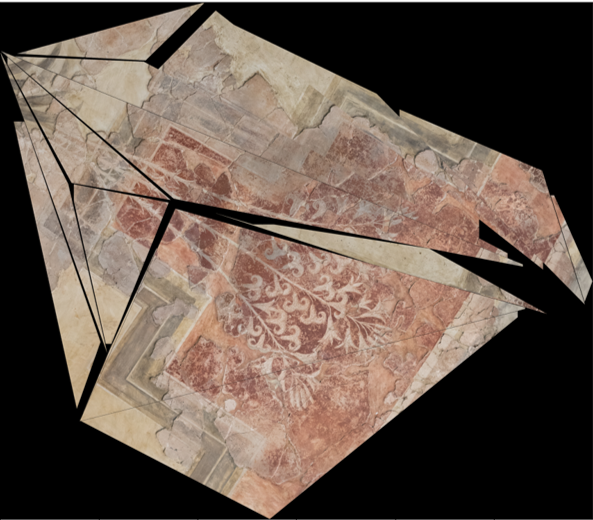} & \includegraphics[height=0.18\textwidth]{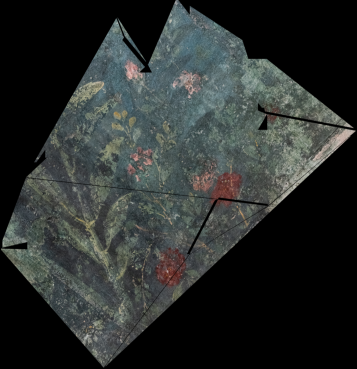} & \includegraphics[height=0.18\textwidth]{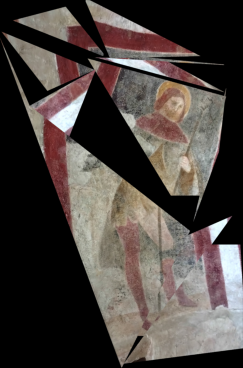}\\
        (A) & (B) & (C) & (D)
    \end{tabular}
    \caption{Selected qualitative results of the convex partition puzzle solver. 
    {\bf (A):} A solution to a CC puzzle~\cite{harel2024pictorial} with both precision and recall being 
    100\%.{\bf (B):} A solution to a Convex Partition puzzle from our new dataset, exhibiting with 95\% precision and 95\% recall.{\bf (C):} A result from our dataset demonstrating perfect precision and 86.36\% recall.{\bf (D):} Another example from our dataset, achieving 93.33\% for both precision and recall.}
    \label{fig:qualatative_puzzle_solving}
\end{figure}

Since Convex partitions are a strict subset of CC puzzles~\cite{harel2024pictorial}, we sought to validate backward compatibility by testing our solver on CC puzzles. 
Since, among other properties, the latter are characterized by cycles of size 4, and since the solver by Harel et al.~\cite{harel2024pictorial} exploited that property explicitly, we anticipated that our more general solver would exhibit slightly inferior performance on such puzzles. 
Indeed, Table~\ref{table:competition} shows this, although in general results are competitive. Importantly, being able to solve CC puzzles demonstrates the backward compatibility of the proposed solver. 
Next to the quantitative results, Fig.~\ref{fig:qualatative_puzzle_solving} shows some qualitative results of our convex partition solver on both convex partition puzzles and on CC puzzles from Harel et al.~\cite{harel2024pictorial}.

\section{Conclusion}

In this work, we introduced a new type of visual puzzle based on Convex Partitions. To our best knowledge, this is the first work to address visual puzzles of that sort, which are a strict generalization of the square, rectangular, and polygonal puzzles studied in the prior art. 
We developed a solver and reported performance on a novel dataset of such puzzles. A unique property in this problem compared to previously studied pictorial puzzles is the unknown number of fragments/pieces that meet at a vertex and thus the need to perform an unconstrained search for cycles in order to reconstruct sub-assemblies. The final solution is then put together by properly merging these cycles into aggregates. Once a full mating graph is hypothesized, a numerical procedure is used to optimally position the noisy pieces as if they are physical fragments connected by springs that seek to minimize their combined potential energy.

Several extensions seem natural for near-future work. In addition to incorporating data-driven methods, this includes the possibility of handling large-scale convex partition puzzles with strong geometric (erosion) noise. We hope that with a broader look at puzzle solving with more general puzzle types, so is the future possibility is to get even closer to solving real-life applications (e.g., archeology~\cite{tsesmelis2024reassembling}).  

\bibliographystyle{plain}   

\begin{thebibliography}{10}

\bibitem{adluru2015sequential}
Nagesh Adluru, Xingwei Yang, and Longin~Jan Latecki.
\newblock Sequential monte carlo for maximum weight subgraphs with application to solving image jigsaw puzzles.
\newblock {\em International journal of computer vision}, 112:319--341, 2015.

\bibitem{alajlan2009solving}
Naif Alajlan.
\newblock Solving square jigsaw puzzles using dynamic programming and the hungarian procedure.
\newblock {\em American Journal of Applied Sciences}, 6(11):1941, 2009.

\bibitem{andalo2016automatic}
Fernanda~A Andal{\'o}, Gustavo Carneiro, Gabriel Taubin, Siome Goldenstein, and Luiz Velho.
\newblock Automatic reconstruction of ancient portuguese tile panels.
\newblock {\em IEEE Comput. Graphics Appl}, 2016.

\bibitem{andalo2012solving}
Fernanda~A Andal{\'o}, Gabriel Taubin, and Sione Goldenstein.
\newblock Solving image puzzles with a simple quadratic programming formulation.
\newblock In {\em 2012 25th SIBGRAPI Conference on Graphics, Patterns and Images}, pages 63--70. IEEE, 2012.

\bibitem{brandao2016hot}
Susana Brand{\~a}o and Manuel Marques.
\newblock Hot tiles: A heat diffusion based descriptor for automatic tile panel assembly.
\newblock In {\em Computer Vision--ECCV 2016 Workshops: Amsterdam, The Netherlands, October 8-10 and 15-16, 2016, Proceedings, Part I 14}, pages 768--782. Springer, 2016.

\bibitem{brown2008system}
Benedict~J Brown, Corey Toler-Franklin, Diego Nehab, Michael Burns, David Dobkin, Andreas Vlachopoulos, Christos Doumas, Szymon Rusinkiewicz, and Tim Weyrich.
\newblock A system for high-volume acquisition and matching of fresco fragments: Reassembling theran wall paintings.
\newblock {\em ACM transactions on graphics (TOG)}, 27(3):1--9, 2008.

\bibitem{Box2D}
Erin Catto.
\newblock Box2d.
\newblock \url{https://box2d.org/}.

\bibitem{cho2010probabilistic}
Taeg~Sang Cho, Shai Avidan, and William~T Freeman.
\newblock A probabilistic image jigsaw puzzle solver.
\newblock In {\em 2010 IEEE Computer society conference on computer vision and pattern recognition}, pages 183--190. IEEE, 2010.

\bibitem{cormen2022introduction}
Thomas~H Cormen, Charles~E Leiserson, Ronald~L Rivest, and Clifford Stein.
\newblock Introduction to algorithms: Fourth editions, 2022.

\bibitem{davis1962norm}
Chandler Davis.
\newblock The norm of the schur product operation.
\newblock {\em Numerische Mathematik}, 4(1):343--344, 1962.

\bibitem{demaine2007jigsaw}
Erik~D. Demaine and Martin~L. Demaine.
\newblock Jigsaw puzzles, edge matching, and polyomino packing: Connections and complexity.
\newblock {\em Graphs and Combinatorics}, 23(1):195--208, 2007.

\bibitem{demaine2020computing}
Erik~D Demaine, S{\'a}ndor~P Fekete, Phillip Keldenich, Dominik Krupke, and Joseph~SB Mitchell.
\newblock Computing convex partitions for point sets in the plane: The cg: shop challenge 2020.
\newblock {\em arXiv preprint arXiv:2004.04207}, 2020.

\bibitem{derech2021solving}
Niv Derech, Ayellet Tal, and Ilan Shimshoni.
\newblock Solving archaeological puzzles.
\newblock {\em Pattern Recognition}, 119:108065, 2021.

\bibitem{dice1945measures}
Lee~R Dice.
\newblock Measures of the amount of ecologic association between species.
\newblock {\em Ecology}, 26(3):297--302, 1945.

\bibitem{elkin2025recognizing}
Gur Elkin, Ofir~Itzhak Shahar, Yaniv Ohayon, Nadav Alali, and Ohad Ben-Shahar.
\newblock Recognizing artistic style of archaeological image fragments using deep style extrapolation.
\newblock In Matthias Rauterberg, editor, {\em Culture and Computing: 13th International Conference, C\&C 2025, Held as Part of the 27th HCI International Conference, HCII 2025, Gothenburg, Sweden, June 22--27, 2025, Proceedings, Part I}, volume 15800 of {\em Lecture Notes in Computer Science}, pages 115--131, Cham, Switzerland, 2025. Springer Cham.

\bibitem{fei2007image}
Ni~Fei, Fu~Zhuang, Liu Renqiang, Cao Qixin, and Zhao Yanzheng.
\newblock An image processing approach for jigsaw puzzle assembly.
\newblock {\em Assembly Automation}, 27(1):25--30, 2007.

\bibitem{freeman1964apictorial}
Herbert Freeman and L~Garder.
\newblock Apictorial jigsaw puzzles: The computer solution of a problem in pattern recognition.
\newblock {\em IEEE Transactions on Electronic Computers}, (2):118--127, 1964.

\bibitem{gallagher2012jigsaw}
Andrew~C Gallagher.
\newblock Jigsaw puzzles with pieces of unknown orientation.
\newblock In {\em 2012 IEEE Conference on computer vision and pattern recognition}, pages 382--389. IEEE, 2012.

\bibitem{giuliari2024positional}
Francesco Giuliari, Gianluca Scarpellini, Stefano Fiorini, Stuart James, Pietro Morerio, Yiming Wang, and Alessio Del~Bue.
\newblock Positional diffusion: Graph-based diffusion models for set ordering.
\newblock {\em Pattern Recognition Letters}, 186:272--278, 2024.

\bibitem{gur2017square}
Shir Gur and Ohad Ben-Shahar.
\newblock From square pieces to brick walls: The next challenge in solving jigsaw puzzles.
\newblock In {\em Proceedings of the IEEE International Conference on Computer Vision}, pages 4029--4037, 2017.

\bibitem{harel2021crossing}
Peleg Harel and Ohad Ben-Shahar.
\newblock Crossing cuts polygonal puzzles: Models and solvers.
\newblock In {\em Proceedings of the IEEE/CVF Conference on Computer Vision and Pattern Recognition}, pages 3084--3093, 2021.

\bibitem{harel2024pictorial}
Peleg Harel, Ofir~Itzhak Shahar, and Ohad Ben-Shahar.
\newblock Pictorial and apictorial polygonal jigsaw puzzles from arbitrary number of crossing cuts.
\newblock {\em International Journal of Computer Vision}, pages 1--35, 2024.

\bibitem{horn2012matrix}
Roger~A Horn and Charles~R Johnson.
\newblock {\em Matrix analysis}.
\newblock Cambridge university press, 2012.

\bibitem{jalkanen2017semi}
Tommi Jalkanen.
\newblock Semi-automatic solving of" jigsaw puzzles" for material re-construction of dead sea scrolls.
\newblock {\em Computer Science}, 28:0, 2017.

\bibitem{jin2023interactively}
Yifan Jin and Xi~Yang.
\newblock Interactively rejioning 2d oracle bone fragments based on contour matching.
\newblock In {\em 2023 9th International Conference on Virtual Reality (ICVR)}, pages 163--170. IEEE, 2023.

\bibitem{khoroshiltseva2021jigsaw}
Marina Khoroshiltseva, Ben Vardi, Alessandro Torcinovich, Arianna Traviglia, Ohad Ben-Shahar, and Marcello Pelillo.
\newblock Jigsaw puzzle solving as a consistent labeling problem.
\newblock In {\em Computer Analysis of Images and Patterns: 19th International Conference, CAIP 2021, Virtual Event, September 28--30, 2021, Proceedings, Part II 19}, pages 392--402. Springer, 2021.

\bibitem{kong2001solving}
Weixin Kong and Benjamin~B Kimia.
\newblock On solving 2d and 3d puzzles using curve matching.
\newblock In {\em Proceedings of the 2001 IEEE Computer Society Conference on Computer Vision and Pattern Recognition. CVPR 2001}, volume~2, pages II--II. IEEE, 2001.

\bibitem{le2019jigsawnet}
Canyu Le and Xin Li.
\newblock Jigsawnet: Shredded image reassembly using convolutional neural network and loop-based composition.
\newblock {\em IEEE Transactions on Image Processing}, 28(8):4000--4015, 2019.

\bibitem{liu2024solving}
Jinyang Liu, Wondmgezahu Teshome, Sandesh Ghimire, Mario Sznaier, and Octavia Camps.
\newblock Solving masked jigsaw puzzles with diffusion vision transformers.
\newblock In {\em Proceedings of the IEEE/CVF Conference on Computer Vision and Pattern Recognition}, pages 23009--23018, 2024.

\bibitem{marande2007mitochondrial}
William Marande and Gertraud Burger.
\newblock Mitochondrial dna as a genomic jigsaw puzzle.
\newblock {\em Science}, 318(5849):415--415, 2007.

\bibitem{mondal2013robust}
Debajyoti Mondal, Yang Wang, and Stephane Durocher.
\newblock Robust solvers for square jigsaw puzzles.
\newblock In {\em 2013 International Conference on Computer and Robot Vision}, pages 249--256. IEEE, 2013.

\bibitem{murakami2008assembly}
Takenori Murakami, Fubito Toyama, Kenji Shoji, and Juichi Miyamichi.
\newblock Assembly of puzzles by connecting between blocks.
\newblock In {\em 2008 19th International Conference on Pattern Recognition}, pages 1--4. IEEE, 2008.

\bibitem{ostertag2020matching}
Cecilia Ostertag and Marie Beurton-Aimar.
\newblock Matching ostraca fragments using a siamese neural network.
\newblock {\em Pattern Recognition Letters}, 131:336--340, 2020.

\bibitem{paikin2015solving}
Genady Paikin and Ayellet Tal.
\newblock Solving multiple square jigsaw puzzles with missing pieces.
\newblock In {\em Proceedings of the IEEE conference on computer vision and pattern recognition}, pages 4832--4839, 2015.

\bibitem{papaodysseus2002contour}
Constantin Papaodysseus, Thanasis Panagopoulos, Michael Exarhos, Constantin Triantafillou, Dimitrios Fragoulis, and Christos Doumas.
\newblock Contour-shape based reconstruction of fragmented, 1600 bc wall paintings.
\newblock {\em IEEE Transactions on Signal Processing}, 50(6):1277--1288, 2002.

\bibitem{pomeranz2011fully}
Dolev Pomeranz, Michal Shemesh, and Ohad Ben-Shahar.
\newblock A fully automated greedy square jigsaw puzzle solver.
\newblock In {\em CVPR 2011}, pages 9--16. IEEE, 2011.

\bibitem{rika2019novel}
Daniel Rika, Dror Sholomon, Eli David, and Nathan~S Netanyahu.
\newblock A novel hybrid scheme using genetic algorithms and deep learning for the reconstruction of portuguese tile panels.
\newblock In {\em Proceedings of the Genetic and Evolutionary Computation Conference}, pages 1319--1327, 2019.

\bibitem{Rombach_2022_CVPR}
Robin Rombach, Andreas Blattmann, Dominik Lorenz, Patrick Esser, and Bj\"orn Ommer.
\newblock High-resolution image synthesis with latent diffusion models.
\newblock In {\em Proceedings of the IEEE/CVF Conference on Computer Vision and Pattern Recognition (CVPR)}, pages 10684--10695, June 2022.

\bibitem{scarpellini2024diffassemble}
Gianluca Scarpellini, Stefano Fiorini, Francesco Giuliari, Pietro Moreiro, and Alessio Del~Bue.
\newblock Diffassemble: A unified graph-diffusion model for 2d and 3d reassembly.
\newblock In {\em Proceedings of the IEEE/CVF Conference on Computer Vision and Pattern Recognition}, pages 28098--28108, 2024.

\bibitem{sholomon2014generalized}
Dror Sholomon, Omid David, and Nathan Netanyahu.
\newblock A generalized genetic algorithm-based solver for very large jigsaw puzzles of complex types.
\newblock In {\em Proceedings of the AAAI Conference on Artificial Intelligence}, volume~28, 2014.

\bibitem{sholomon2013genetic}
Dror Sholomon, Omid David, and Nathan~S Netanyahu.
\newblock A genetic algorithm-based solver for very large jigsaw puzzles.
\newblock In {\em Proceedings of the IEEE conference on computer vision and pattern recognition}, pages 1767--1774, 2013.

\bibitem{sizikova2017wall}
Elena Sizikova and Thomas Funkhouser.
\newblock Wall painting reconstruction using a genetic algorithm.
\newblock {\em Journal on Computing and Cultural Heritage (JOCCH)}, 11(1):1--17, 2017.

\bibitem{son2014solving}
Kilho Son, James Hays, and David~B Cooper.
\newblock Solving square jigsaw puzzles with loop constraints.
\newblock In {\em Computer Vision--ECCV 2014: 13th European Conference, Zurich, Switzerland, September 6-12, 2014, Proceedings, Part VI 13}, pages 32--46. Springer, 2014.

\bibitem{son2018solving}
Kilho Son, James Hays, and David~B Cooper.
\newblock Solving square jigsaw puzzle by hierarchical loop constraints.
\newblock {\em IEEE transactions on pattern analysis and machine intelligence}, 41(9):2222--2235, 2018.

\bibitem{son2016solving}
Kilho Son, James Hays, David~B Cooper, et~al.
\newblock Solving small-piece jigsaw puzzles by growing consensus.
\newblock In {\em Proceedings of the IEEE Conference on Computer Vision and Pattern Recognition}, pages 1193--1201, 2016.

\bibitem{toyama2002assembly}
Fubito Toyama, Yukihiro Fujiki, Kenji Shoji, and Juichi Miyamichi.
\newblock Assembly of puzzles using a genetic algorithm.
\newblock In {\em 2002 International Conference on Pattern Recognition}, volume~4, pages 389--392. IEEE, 2002.

\bibitem{tsesmelis2024reassembling}
Theodore Tsesmelis, Luca Palmieri, Marina Khoroshiltseva, Adeela Islam, Gur Elkin, Ofir~Itzhak Shahar, Gianluca Scarpellini, Stefano Fiorini, Yaniv Ohayon, Nadav Alali, et~al.
\newblock Re-assembling the past: The repair dataset and benchmark for real world 2d and 3d puzzle solving.
\newblock In {\em NeurIPS}, 2024.

\bibitem{vardi2023multi}
Ben Vardi, Alessandro Torcinovich, Marina Khoroshiltseva, Marcello Pelillo, and Ohad Ben-Shahar.
\newblock Multi-phase relaxation labeling for square jigsaw puzzle solving.
\newblock {\em arXiv preprint arXiv:2303.14793}, 2023.

\bibitem{yang2011particle}
Xingwei Yang, Nagesh Adluru, and Longin~Jan Latecki.
\newblock Particle filter with state permutations for solving image jigsaw puzzles.
\newblock In {\em CVPR 2011}, pages 2873--2880. IEEE, 2011.

\bibitem{yu2015solving}
Rui Yu, Chris Russell, and Lourdes Agapito.
\newblock Solving jigsaw puzzles with linear programming.
\newblock {\em arXiv preprint arXiv:1511.04472}, 2015.

\bibitem{zhao2007puzzle}
Yu-Xiang Zhao, Mu-Chun Su, Zhong-Lie Chou, and Jonathan Lee.
\newblock A puzzle solver and its application in speech descrambling.
\newblock In {\em WSEAS International Conference on Computer Engineering and Applications}, pages 171--176, 2007.

\bibitem{zheng2024reunion}
Yutong Zheng, Xuelong Li, and Yu~Weng.
\newblock Reunion helper: an edge matcher for sibling fragment identification of the dunhuang manuscript.
\newblock {\em Heritage Science}, 12(1):52, 2024.

\bibitem{zhou2025pairingnet}
Rixin Zhou, Ding Xia, Yi~Zhang, Honglin Pang, Xi~Yang, and Chuntao Li.
\newblock Pairingnet: A learning-based pair-searching and-matching network for image fragments.
\newblock In {\em European Conference on Computer Vision}, pages 234--251. Springer, 2025.


\bibitem{itzhak2025pairwise}
Ofir~Itzhak Shahar, Gur~Elkin, and Ohad~Ben-Shahar.
\newblock Pairwise alignment \& compatibility for arbitrarily irregular image fragments.
\newblock {\em arXiv e-prints}, pages arXiv--2507, 2025.


\end{thebibliography}

\end{document}